\documentclass{article}
\usepackage{nips14submit_e,times}
\usepackage{algorithm}
\usepackage{algorithmicx}
\usepackage{makeidx}
\usepackage{url}
\usepackage{bm}
\usepackage{lscape}
\usepackage{latexsym}
\usepackage{algpseudocode}
\usepackage{tabularx}
\usepackage{amsmath}
\usepackage{amsfonts}
\usepackage{amssymb}
\usepackage[normalem]{ulem}
\usepackage{epsfig}

\newtheorem{theorem}{Theorem}
\newtheorem{lemma}[theorem]{Lemma}
\newtheorem{proposition}[theorem]{Proposition}

\newtheorem{corollary}[theorem]{Corollary}
\newtheorem{definition}[theorem]{Definition} 

\newcommand{\argmin}{\operatornamewithlimits{argmin}}

\newcommand{\BlackBox}{\rule{1.5ex}{1.5ex}}  
\newenvironment{proof}{\par\noindent{\bf Proof\ \ \ }}{\hfill\BlackBox\\[2mm]}
\def\endproof{\hfill$\sqcap \!\!\!\! \sqcup$\bigskip}  

\title{Structure Regularization for Structured Prediction: Theories and Experiments}

\author{
Xu Sun$^{* \dag}$ \\
$*$MOE Key Laboratory of Computational Linguistics, Peking University \\
$\dag$School of Electronics Engineering and Computer Science, Peking University \\
\texttt{xusun@pku.edu.cn} \\
}

\nipsfinalcopy 

\begin{document}

\maketitle

\begin{abstract}
While there are many studies on weight regularization, the study on structure regularization is rare.
Many existing systems on structured prediction focus on increasing the level of structural dependencies within the model.
However, this trend could have been misdirected, because our study suggests that complex structures are actually harmful to generalization ability in structured prediction. To control structure-based overfitting, we propose a structure regularization framework via \emph{structure decomposition}, which decomposes training samples into mini-samples with simpler structures, deriving a model with better generalization power. We show both theoretically and empirically that structure regularization can effectively control overfitting risk and lead to better accuracy. As a by-product, the proposed method can also substantially accelerate the training speed. The method and the theoretical results can apply to general graphical models with arbitrary structures. Experiments on well-known tasks demonstrate that our method can easily beat the benchmark systems on those highly-competitive tasks, achieving state-of-the-art accuracies yet with substantially faster training speed.
\end{abstract}

\section{Introduction}

Structured prediction models are popularly used to solve structure dependent problems in a wide variety of application domains including natural language processing, bioinformatics, speech recognition, and computer vision.
To solve those problems, many structured prediction methods have been developed, with representative models such as conditional random fields (CRFs), deep neural networks, and structured perceptron models.
Recently, in order to more accurately capture structural information, some studies emphasize on intensifying structural dependencies in structured prediction, such as applying long range dependencies among tags and developing long distance features or global features.

We argue that over-emphasis on intensive structural dependencies could be misleading, because our study suggests that complex structures are actually harmful to model accuracy. Indeed, while it is obvious that intensive structural dependencies can effectively incorporate structural information, it is less obvious that intensive structural dependencies have a drawback of increasing the generalization risk. Increasing the generalization risk means the trained model tends to overfit the training data, because more complex structures are easier to suffer from overfitting. Formally, our theoretical analysis reveals why and with what degree the structure complexity lowers the generalization ability of trained models.
Since this type of overfitting is caused by structure complexity, it can hardly be solved by ordinary regularization methods such as $L_2$ and $L_1$ regularization schemes, which is only for controlling weight complexity.

To deal with this problem, we propose a simple structure regularization solution based on \emph{tag structure decomposition}. The proposed method decomposes each training sample into multiple mini-samples with simpler structures, deriving a model with better generalization power. The proposed method is easy to implement, and it has several interesting properties: (1) We show both theoretically and empirically that the proposed method can reduce the overfit risk. (2) Keeping the convexity of the objective function: a convex function with a structure regularizer is still convex. (3) No conflict with the weight regularization: we can apply structure regularization together with weight regularization. (4) Accelerating the convergence rate in training. (5) This method can be used for different types of models, including CRFs \cite{LaffertyMcCallum01} and perceptrons \cite{Collins2002}.

The term \emph{structural regularization} has been used in prior work for regularizing \emph{structures of features}. For (typically non-structured) classification problems, there are considerable studies on structure-related regularization, including spectral regularization for modeling feature structures in multi-task learning \cite{Argyriou07}, regularizing feature structures for structural large margin classifiers \cite{tnn/XueCY11}, and many recent studies on structured sparsity. Structure sparsity is studied for a variety of non-structured classification models \cite{nips/MicchelliMP10,icml/DuchiS09} and structured prediction scenarios \cite{jmlr/SchmidtM10,MartinsSFA11a}, via adopting mixed norm regularization \cite{icml/QuattoniCCD09}, \emph{Group Lasso} \cite{Yuan06mod}, posterior regularization \cite{nips/GracaGTP09}, and a string of variations \cite{corr/Bach2011,sac/ObozinskiTJ10,jmlr/HuangZM11}.
Compared with those prior work, we emphasize that our proposal on tag structure regularization is novel. This is because the term \emph{structure} in all of the aforementioned work refers to \emph{structures of feature space}, which is substantially different compared with our proposal on regularizing tag structures (interactions among tags).

There are other related studies, including the studies of \cite{icml/SuttonM07} and \cite{conf/icml/SamdaniR12} on piecewise/decomposed training methods, and the study of \cite{Tsuruoka2011} on a ``lookahead" learning method. Our work differs from \cite{icml/SuttonM07,conf/icml/SamdaniR12,Tsuruoka2011} mainly because our work is built on a regularization framework, with arguments and justifications on reducing generalization risk and for better accuracy. Also, our method and the theoretical results can fit general graphical models with arbitrary structures, and the detailed algorithm is quite different.
On generalization risk analysis, related studies include \cite{jmlr/BousquetE02,colt/ShwartzSSS09a} on non-structured classification and \cite{Taskar2003,London2013,london2013collective} on structured classification.

To the best of our knowledge, this is the first theoretical result on quantifying the relation between structure complexity and the generalization risk in structured prediction, and this is also the first proposal on structure regularization via regularizing tag-interactions.
The contributions of this work\footnote{See the code at \url{http://klcl.pku.edu.cn/member/sunxu/code.htm}} are two-fold:
\begin{itemize}
\item On the methodology side, we propose a general purpose structure regularization framework for structured prediction. We show both theoretically and empirically that the proposed method can effectively reduce the overfitting risk in structured prediction, and that the proposed method also has an interesting by-product of accelerating the rates of convergence in training. The structure regularization method and the theoretical analysis do \emph{not} make assumptions or constraints based on specific structures. In other words, the method and the theoretical results can apply to graphical models with arbitrary structures, including linear chains, trees, and general graphs.

\item On the application side, for several important natural language processing tasks, including part-of-speech tagging, biomedical entity recognition, and word segmentation, our simple method can easily beat the benchmark systems on those highly-competitive tasks, achieving record-breaking accuracies as well as substantially faster training speed.
\end{itemize}

\section{Structure Regularization}

We first describe the proposed structure regularization method, and then give theoretical results on analyzing generalization risk and convergence rates.

\subsection{Settings}\label{sec.intro}

A graph of observations (even with arbitrary structures) can be indexed and be denoted by using an indexed sequence of observations $\pmb O=\{o_1, \dots, o_n\}$.
We use the term \emph{sample} to denote $\pmb O=\{o_1, \dots, o_n\}$.
For example, in natural language processing, a sample may correspond to a sentence of $n$ words with dependencies of linear chain structures (e.g., in part-of-speech tagging) or tree structures (e.g., in syntactic parsing). In signal processing, a sample may correspond to a sequence of $n$ signals with dependencies of arbitrary structures. For simplicity in analysis, we assume all samples have $n$ observations (thus $n$ tags). In a typical setting of structured prediction, all the $n$ tags have inter-dependencies via connecting each Markov dependency between neighboring tags. Thus, we call $n$ as \emph{tag structure complexity} or simply \emph{structure complexity} below.

A sample is converted to an indexed sequence of feature vectors $\pmb x =\{\pmb x_{(1)}, \dots, \pmb x_{(n)}\}$, where $\pmb x_{(k)} \in \mathcal X$ is of the dimension $d$ and corresponds to the local features extracted from the position/index $k$.\footnote{In most of the existing structured prediction methods, including conditional random fields (CRFs), all the local feature vectors should have the same dimension of features.} We can use an $n \times d$ matrix to represent $\pmb x \in \mathcal X^n$. In other words, we use $\mathcal X$ to denote the input space on a position, so that $\pmb x$ is sampled from $\mathcal X^n$.
Let $\mathcal Y^n \subset \mathbb R^n$ be structured output space, so that the structured output $\pmb y$ are sampled from $\mathcal Y^n$.
Let $\mathcal Z=(\mathcal X^n,\mathcal Y^n)$ be a unified denotation of structured input and output space. Let $\pmb z = (\pmb x, \pmb y)$, which is sampled from $\mathcal Z$, be a unified denotation of a $(\pmb x, \pmb y)$ pair in the training data.

Suppose a training set is
$$
S=\{\pmb z_1=(\pmb x_1, \pmb y_1), \dots, \pmb z_m=(\pmb x_m, \pmb y_m) \},
$$
with size $m$, and the samples are drawn i.i.d. from a distribution $D$ which is unknown. A learning algorithm is a function $G: \mathcal Z^m \mapsto \mathcal F$ with the function space $\mathcal F \subset  \{\mathcal X^n \mapsto \mathcal Y^n\}$, i.e., $G$ maps a training set $S$ to a function $G_S: \mathcal X^n \mapsto \mathcal Y^n $.
We suppose $G$ is symmetric with respect to $S$, so that $G$ is independent on the order of $S$.

Structural dependencies among tags are the major difference between structured prediction and non-structured classification. For the latter case, a local classification of $g$ based on a position $k$ can be expressed as $g(\pmb x_{(k-a)}, \dots, \pmb x_{(k+a)})$, where the term $\{\pmb x_{(k-a)}, \dots, \pmb x_{(k+a)}\}$ represents a local window.
However, for structured prediction, a local classification on a position depends on the whole input $\pmb x =\{\pmb x_{(1)}, \dots, \pmb x_{(n)}\}$ rather than a local window, due to the nature of structural dependencies among tags (e.g., graphical models like CRFs). Thus, in structured prediction a local classification on $k$ should be denoted as $g(\pmb x_{(1)}, \dots, \pmb x_{(n)}, k)$. To simplify the notation, we define
$$g(\pmb x, k) \triangleq g(\pmb x_{(1)}, \dots, \pmb x_{(n)}, k)$$

Given a training set $S$ of size $m$, we define $S^{\setminus i}$ as a modified training set, which removes the $i$'th training sample:
$$
S^{\setminus i}=\{ \pmb z_1, \dots, \pmb z_{i-1}, \pmb z_{i+1}, \dots, \pmb z_m \},
$$
and we define $S^i$ as another modified training set, which replaces the $i$'th training sample with a new sample $\pmb{\hat z}_i$ drawn from $D$:
$$
S^i=\{ \pmb z_1, \dots, \pmb z_{i-1}, \pmb {\hat z}_i, \pmb z_{i+1}, \dots, \pmb z_m \},
$$

We define \emph{point-wise cost function} $c: \mathcal Y \times \mathcal Y \mapsto \mathbb R^+$ as $c[G_S(\pmb x, k), \pmb y_{(k)}]$, which measures the cost on a position $k$ by comparing $G_S(\pmb x, k)$ and the gold-standard tag $\pmb y_{(k)}$, and we introduce the point-wise loss as
$$
\ell (G_S, \pmb z, k) \triangleq c[G_S(\pmb x, k), \pmb y_{(k)}]
$$

Then, we define \emph{sample-wise cost function} $C: \mathcal Y^n \times \mathcal Y^n \mapsto \mathbb R^+$, which is the cost function with respect to a whole sample, and we introduce the sample-wise loss as
$$
\mathcal L(G_S, \pmb z)
\triangleq C[G_S(\pmb x), \pmb y]
= \sum_{k=1}^n \ell (G_S, \pmb z, k)
= \sum_{k=1}^n c[G_S(\pmb x, k), \pmb y_{(k)}]
$$

Given $G$ and a training set $S$, what we are most interested in is the \emph{generalization risk} in structured prediction (i.e., expected average loss) \cite{Taskar2003,London2013}:
$$
R(G_S)= \mathbb E_{\pmb z} \Big[ \frac {\mathcal L(G_S, \pmb z)} n \Big] 
$$

Unless specifically indicated in the context, the probabilities and expectations over random variables, including $\mathbb E_{\pmb z}(.)$, $\mathbb E_S(.)$, $\mathbb P_{\pmb z}(.)$, and $\mathbb P_S(.)$, are based on the unknown distribution $D$.

Since the distribution $D$ is unknown, we have to estimate $R(G_S)$ from $S$ by  using the \emph{empirical risk}:
$$
R_e (G_S)= \frac 1 {mn} \sum_{i=1}^m \mathcal L(G_S,\pmb z_i)
= \frac 1 {mn} \sum_{i=1}^m \sum_{k=1}^n \ell (G_S, \pmb z_i, k)
$$

In what follows, sometimes we will use simplified notations, $R$ and $R_e$, to denote $R(G_S)$ and $R_e(G_S)$.

To state our theoretical results, we must describe several quantities and assumptions which are important in structured prediction.
We follow some notations and assumptions on non-structured classification \cite{jmlr/BousquetE02,colt/ShwartzSSS09a}.
We assume a simple real-valued structured prediction scheme such that the class predicted on position $k$ of $\pmb{x}$ is the sign of $G_S(\pmb x, k)\in \mathcal D$.\footnote{In practice, many popular structured prediction models have a convex and real-valued cost function (e.g., CRFs).} Also, we assume the point-wise cost function $c_\tau$ is convex and \emph{$\tau$-smooth} such that $\forall y_1, y_2 \in \mathcal D, \forall y^* \in \mathcal Y$
\begin{equation}\label{eq12}
|c_\tau(y_1, y^*) - c_\tau(y_2,y^*)| \leq \tau|y_1 - y_2|
\end{equation}

Then, \emph{$\tau$-smooth} versions of the loss and the cost function can be derived according to their prior definitions:
$$
\mathcal L_\tau (G_S, \pmb z) = C_\tau [G_S(\pmb x), \pmb y]
= \sum_{k=1}^n \ell_\tau (G_S,\pmb z, k) =  \sum_{k=1}^n c_\tau [G_S(\pmb x, k), \pmb y_{(k)}]
$$

Also, we use a value $\rho$ to quantify the bound of $|G_S(\pmb x, k) - G_{S^{\setminus i}}(\pmb x, k)|$ while changing a single sample (with size $n' \leq n$) in the training set
with respect to the structured input $\pmb x$. This $\rho$\emph{-admissible} assumption can be formulated as $\forall k$,
\begin{equation}\label{eq14}
|G_S(\pmb x, k) - G_{S^{\setminus i}}(\pmb x, k)| \leq \rho||G_S-G_{S^{\setminus i}}||_2 \cdot ||\pmb x||_2
\end{equation}
where $\rho \in \mathbb R^+$ is a value related to the design of algorithm $G$.

\subsection{Structure Regularization}

\begin{figure}[t]
\begin{center}
\begin{tabular}{c}
    \includegraphics[width=1\hsize]{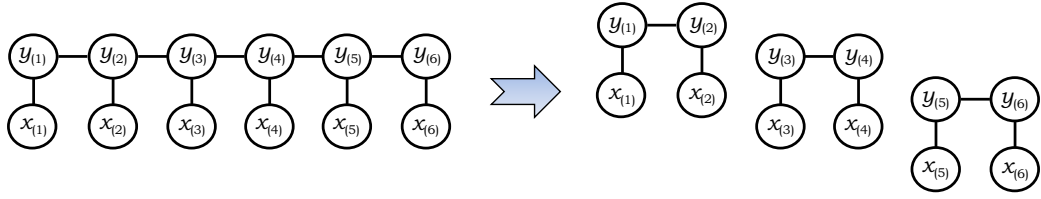}
\end{tabular}
\caption{An illustration of structure regularization in simple linear chain case, which decompose a training sample $\pmb z$ with structure complexity 6 into three mini-samples with structure complexity 2. Structure regularization can apply to more general graphs with arbitrary dependencies.
}\label{fig.example}
\end{center}
\end{figure}

\begin{algorithm}[tb]
   \caption{Training with structure regularization}
   \label{algo:sr}
\begin{algorithmic}[1]

 \State {\textbf{Input}: model weights $\pmb w$, training set $S$, structure regularization strength $\alpha$}
 \Repeat
 \State {$S' \gets \emptyset$}
 \For {$i = 1 \to m$}
 \State {Randomly decompose $\pmb z_i \in S$ into mini-samples $N_\alpha(\pmb z_i)=\{\pmb z_{(i,1)}, \dots, \pmb z_{(i,\alpha)}\}$}
 \State {$S' \gets S' \cup N_\alpha(\pmb z_i)$}
 \EndFor

 \For {$i = 1 \to |S'|$}
 \State {Sample $\pmb z'$ uniformly at random from $S'$, with gradient $\nabla g_{\pmb z'}(\pmb w)$}
 \State {$\pmb w \gets \pmb w -\eta \nabla g_{\pmb z'}(\pmb w)$}
 \EndFor
 \Until {Convergence}
 \State \Return {$\pmb w$}

\end{algorithmic}
\end{algorithm}

Most existing regularization techniques are for regularizing model weights/parameters (e.g., a representative regularizer is the Gaussian regularizer or so called $L_2$ regularizer), and we call such regularization techniques as \emph{weight regularization}.
\begin{definition}[Weight regularization]\label{def1}
Let $N_\lambda: \mathcal F \mapsto \mathbb R^+$ be a weight regularization function on $\mathcal F$ with regularization strength $\lambda$, the structured classification based objective function with general weight regularization is as follows:
\begin{equation}
R_{\lambda}(G_S) \triangleq R_e(G_S) + N_\lambda(G_S)
\end{equation}
\end{definition}

While weight regularization is normalizing model weights, the proposed structure regularization method is normalizing the structural complexity of the training samples. As illustrated in Figure~\ref{fig.example}, our proposal is based on \emph{tag structure decomposition}, which can be formally defined as follows:
\begin{definition}[Structure regularization]\label{def2}
Let $N_\alpha: \mathcal F \mapsto \mathcal F$ be a structure regularization function on $\mathcal F$ with regularization strength $\alpha$ with $1\leq \alpha \leq n$, the structured classification based objective function with structure regularization is as follows\footnote{The notation $N$ is overloaded here. For clarity throughout, $N$ with subscript $\lambda$ refers to weight regularization function, and $N$ with subscript $\alpha$ refers to structure regularization function.}:
\begin{equation}
R_{\alpha}(G_S)\triangleq R_e[G_{N_\alpha (S)}]
= \frac 1 {m n} \sum_{i=1}^{m} \sum_{j=1}^\alpha \mathcal L[G_{S'}, \pmb z_{(i,j)}]
= \frac 1 {m n}  \sum_{i=1}^{m} \sum_{j=1}^\alpha \sum_{k=1}^{n/\alpha} \mathcal \ell [G_{S'}, \pmb z_{(i,j)}, k]
\end{equation}
where $N_\alpha(\pmb z_i)$ randomly splits $\pmb z_i$ into $\alpha$ mini-samples $\{\pmb z_{(i,1)}, \dots, \pmb z_{(i,\alpha)}\}$, so that the mini-samples have a distribution on their sizes (structure complexities) with the expected value $n'= n/\alpha$. Thus, we get
\begin{equation}
S'=\{\underbrace{\pmb z_{(1,1)},z_{(1,2)},\dots,\pmb z_{(1,\alpha)}}_{\alpha},\dots,\underbrace{\pmb z_{(m,1)},\pmb z_{(m,2)},\dots,\pmb z_{(m,\alpha)}}_{\alpha} \}
\end{equation}
with $m\alpha$ mini-samples with expected structure complexity $n/\alpha$. We can denote $S'$ more compactly as
$
S'=\{\pmb z_1', \pmb z_2', \dots, \pmb z_{m\alpha}' \}
$
and $R_{\alpha}(G_S)$ can be simplified as
\begin{equation}
R_{\alpha}(G_S)
\triangleq \frac 1 {mn} \sum_{i=1}^{m\alpha} \mathcal L(G_{S'}, \pmb z_i')
= \frac 1 {m n}  \sum_{i=1}^{m\alpha} \sum_{k=1}^{n/\alpha} \mathcal \ell [G_{S'}, \pmb z_i', k]
\end{equation}
\end{definition}
Note that, when the structure regularization strength $\alpha=1$, we have $S'=S$ and $R_{\alpha}=R_e$.
The structure regularization algorithm (with the stochastic gradient descent setting) is summarized in Algorithm \ref{algo:sr}.

Since we know $\pmb z = (\pmb x, \pmb y)$, the decomposition of $\pmb z$ simply means the decomposition of $\pmb x$ and $\pmb y$. Recall that $\pmb x =\{\pmb x_{(1)}, \dots, \pmb x_{(n)}\}$ is an indexed sequence of the feature vectors, not the observations $\pmb O=\{o_1, \dots, o_n\}$. Thus, it should be emphasized that the decomposition of $\pmb x$ is the decomposition of the feature vectors, not the original observations. Actually the decomposition of the feature vectors is more convenient and has no information loss \---- no need to regenerate features. On the other hand, decomposing observations needs to regenerate features and may lose some features.

The structure regularization has no conflict with the weight regularization, and the structure regularization can be applied together with the weight regularization. Actually we will show that applying the structure regularization over the weight regularization can further improve stability and reduce generalization risk.
\begin{definition}[Structure \& weight regularization]
By combining structure regularization in Definition \ref{def2} and weight regularization in Definition \ref{def1},
the structured classification based objective function is as follows:
\begin{equation}
R_{\alpha,\lambda}(G_S)
\triangleq R_{\alpha}(G_S) + N_\lambda(G_S)
\end{equation}
When $\alpha=1$, we have $R_{\alpha,\lambda} =R_e(G_S) + N_\lambda(G_S) =R_\lambda$.
\end{definition}

Like existing weight regularization methods, currently our structure regularization is only for the training stage. Currently we do not use structure regularization in the test stage.

\subsection{Stability of Structured Prediction}

In contrast to the simplicity of the algorithm, the theoretical analysis is quite technical. First, we analyze the stability of structured prediction.

\begin{definition}[Function stability]
A real-valued structured classification algorithm
$G$ has ``function value based stability" (``function stability" for short) $\Delta$ if the following holds: $\forall \pmb{z}=(\pmb x, \pmb y) \in \mathcal Z, \forall S \in \mathcal Z^m, \forall i \in \{1, \dots, m\}, \forall k \in \{1, \dots, n\}$,
\begin{equation*}
|   G_S(\pmb x, k) - G_{S^{\setminus i}}(\pmb x, k) | \leq \Delta
\end{equation*}
\end{definition}

\begin{definition}[Loss stability]
A structured classification algorithm $G$ has ``uniform loss-based stability" (``loss stability'' for short) $\Delta_l$ if the following holds: $\forall \pmb{z} \in \mathcal{Z}, \forall S \in \mathcal{Z}^m, \forall i \in \{1, \dots, m\}, \forall k \in \{1, \dots, n\}$,
\begin{equation*}
|\ell(G_S,\pmb{z},k) - \ell(G_{S^{\setminus i}},\pmb{z},k)  | \leq \Delta_l
\end{equation*}

$G$ has ``sample-wise uniform loss-based stability" (``sample loss stability" for short) $\Delta_s$ with respect to the loss function $\mathcal L$ if the following holds: $\forall \pmb{z} \in \mathcal{Z}, \forall S \in \mathcal{Z}^m, \forall i \in \{1, \dots, m\}$,
\begin{equation*}
|\mathcal{L}(G_S,\pmb{z}) - \mathcal{L}(G_{S^{\setminus i}},\pmb{z})  | \leq \Delta_s
\end{equation*}
\end{definition}

\begin{lemma}[Loss stability vs. function stability]\label{lemma2}
If a real-valued structured classification algorithm $G$ has function stability $\Delta$ with respect to loss function $\ell_\tau$, then $G$ has loss stability
$$\Delta_l \leq \tau \Delta$$
and sample loss stability
$$\Delta_s \leq n \tau \Delta.$$
\end{lemma}

The proof is in Section \ref{proof}.

Here, we show that our structure regularizer can further improve stability (thus reduce generalization risk) over a model which already equipped with a weight regularizer.

\begin{theorem}[Stability vs. structure regularization]\label{theo1.2}
With a training set $S$ of size $m$, let the learning algorithm $G$ have the minimizer $f$ based on commonly used $L_2$ weight regularization:
\begin{equation}\label{eq22}
f = \argmin_{g \in \mathcal F} R_{\alpha,\lambda}(g)
= \argmin_{g \in \mathcal F} \Big( \frac 1 {mn} \sum_{j=1}^{m\alpha} \mathcal L_\tau(g, \pmb z_j') + \frac \lambda {2} ||g||_2^2 \Big)
\end{equation}
where $\alpha$ denotes structure regularization strength with $1\leq \alpha \leq n$.

Also, we have
\begin{equation}\label{eq22.2}
f^{\setminus {i'}} = \argmin_{g \in \mathcal F} R_{\alpha,\lambda}^{\setminus {i'}}(g)
=\argmin_{g \in \mathcal F} \Big( \frac 1 {mn} \sum_{j\neq i'} \mathcal L_\tau(g, \pmb z_j') + \frac \lambda {2} ||g||_2^2 \Big)
\end{equation}
where $j\neq i'$ means $j\in \{1,\dots,i'-1,i'+1,\dots,m\alpha\}$.\footnote{Note that, in some cases the notation $i$ is ambiguous. For example, $f^{\setminus i}$ can either denote the removing of a sample in $S$ or denote the removing of a mini-sample in $S'$. Thus, when the case is ambiguous, we use different index symbols for $S$ and $S'$, with $i$ for indexing $S$ and $i'$ for indexing $S'$, respectively.}
Assume $\mathcal L_\tau$ is convex and differentiable, and $f(\pmb x, k)$ is $\rho$-admissible.
Let a local feature value is bounded by $v$ such that $\pmb x_{(k,q)} \leq v$ for $q \in \{1, \dots, d\}$.\footnote{Recall that $d$ is the dimension of local feature vectors defined in Section~\ref{sec.intro}. }
Let $\Delta$ denote the function stability of $f$ comparing with $f^{\setminus {i'}}$ for $\forall \pmb z \in \mathcal Z$ with $|\pmb z|=n$.
Then, $\Delta$ is bounded by
\begin{equation}\label{eq11}
\Delta \leq \frac {d \tau \rho^2 v^2 n^2} {m\lambda\alpha^2} ,
\end{equation}
and the corresponding loss stability is bounded by $$\Delta_l \leq \frac {d \tau^2 \rho^2 v^2 n^2} {m\lambda\alpha^2},$$
and the corresponding sample loss stability is bounded by $$\Delta_s \leq \frac {d \tau^2 \rho^2 v^2 n^3} {m\lambda\alpha^2}.$$
\end{theorem}

The proof is in Section \ref{proof}.

We can see that increasing the size of training set $m$ results in linear improvement of $\Delta$, and increasing the strength of structure regularization $\alpha$ results in quadratic improvement of $\Delta$.

The function stability $\Delta$ is based on comparing $f$ and $f^{\setminus i'}$, i.e., the stability is based on removing a mini-sample. Moreover, we can extend the analysis to the function stability based on comparing $f$ and $f^{\setminus i}$, i.e., the stability is based on removing a full-size sample.

\begin{corollary}[Stability based on $\setminus i$ rather than $\setminus {i'}$]\label{coro1}
With a training set $S$ of size $m$, let the learning algorithm $G$ have the minimizer $f$ as defined like before.
Also, we have
\begin{equation}\label{eq22.3}
f^{\setminus {i}} = \argmin_{g \in \mathcal F} R_{\alpha,\lambda}^{\setminus {i}}(g)
=\argmin_{g \in \mathcal F} \Big( \frac 1 {mn} \sum_{j\notin i} \mathcal L_\tau(g, \pmb z_j') + \frac \lambda {2} ||g||_2^2 \Big)
\end{equation}
where $j\notin i$ means $j\in \{1,\dots,(i-1)\alpha,i\alpha+1,\dots,m\alpha\}$, i.e., all the mini-samples derived from the sample $\pmb z_i$ are removed.
Assume $\mathcal L_\tau$ is convex and differentiable, and $f(\pmb x, k)$ is $\rho$-admissible.
Let a local feature value is bounded by $v$ such that $\pmb x_{(k,q)} \leq v$ for $q \in \{1, \dots, d\}$.
Let $\bar \Delta$ denote the function stability of $f$ comparing with $f^{\setminus i}$ for $\forall \pmb z \in \mathcal Z$ with $|\pmb z|=n$.
Then, $\bar \Delta$ is bounded by
\begin{equation}\label{eq11.2}
\bar \Delta \leq \frac {d \tau \rho^2 v^2 n^2} {m\lambda\alpha} = \alpha \sup(\Delta) ,
\end{equation}
where $\Delta$ is the function stability of $f$ comparing with $f^{\setminus i'}$, and $\sup(\Delta) =\frac {d \tau \rho^2 v^2 n^2} {m\lambda\alpha^2}$, as described in Eq. (\ref{eq11}). Similarly, we have
$$\bar \Delta_l \leq \frac {d \tau^2 \rho^2 v^2 n^2} {m\lambda\alpha} = \alpha \sup(\Delta_l) ,$$
and
$$\bar \Delta_s \leq \frac {d \tau^2 \rho^2 v^2 n^3} {m\lambda\alpha} = \alpha \sup(\Delta_s) .$$
\end{corollary}

The proof is in Section \ref{proof}.

\subsection{Reduction of Generalization Risk}

\begin{theorem}[Generalization vs. stability]\label{theo2}
Let $G$ be a real-valued structured classification algorithm with a point-wise loss function $\ell_\tau$ such that $\forall k, 0 \leq \ell_\tau (G_S, \pmb z, k) \leq \gamma$. Let $f$, $\Delta$, and $\bar \Delta$ be defined like before. Let $R(f)$ be the generalization risk of $f$ based on the expected sample $\pmb z \in \mathcal Z$ with size $n$, as defined like before. Let $R_e(f)$ be the empirical risk of $f$ based on $S$, as defined like before. Then, for any $\delta \in (0,1)$, with probability at least $1-\delta$ over the random draw of the training set $S$, the generalization risk $R(f)$ is bounded by
\begin{equation}\label{eq10}
R(f) \leq R_e(f) + {2\tau \bar\Delta}  + \Big({(4m-2)\tau \bar\Delta}  + \gamma \Big) \sqrt{\frac {\ln {\delta^{-1}}} {2m}}
\end{equation}
\end{theorem}

The proof is in Section \ref{proof}.

\begin{theorem}[Generalization vs. structure regularization]\label{theo3}
Let the structured prediction objective function of $G$ be penalized by structure regularization with factor $\alpha \in [1,n]$ and $L_2$ weight regularization with factor $\lambda$, and the penalized function has a minimizer $f$:
\begin{equation}
f = \argmin_{g \in \mathcal F} R_{\alpha,\lambda}(g)
= \argmin_{g \in \mathcal F} \Big( \frac 1 {mn} \sum_{j=1}^{m\alpha} \mathcal L_\tau(g, \pmb z_j') + \frac \lambda {2} ||g||_2^2 \Big)
\end{equation}
Assume the point-wise loss $\ell_\tau$ is convex and differentiable, and is bounded by $\ell_\tau (f, \pmb z, k) \leq \gamma$. Assume $f(\pmb x, k)$ is $\rho$-admissible.
Let a local feature value be bounded by $v$ such that $\pmb x_{(k,q)} \leq v$ for $q \in \{1, \dots, d\}$.
Then, for any $\delta \in (0,1)$, with probability at least $1-\delta$ over the random draw of the training set $S$, the generalization risk $R(f)$ is bounded by
\begin{equation}\label{eq23}
R(f) \leq R_e(f) + {\frac {2d \tau^2 \rho^2 v^2 n^2} {m\lambda\alpha}}  + \Big({\frac {(4m-2) d \tau^2 \rho^2 v^2 n^2} {m\lambda\alpha}}  + \gamma \Big) \sqrt{\frac {\ln {\delta^{-1}}} {2m}}
\end{equation}
\end{theorem}

The proof is in Section \ref{proof}.

We call the term ${\frac {2d \tau^2 \rho^2 v^2 n^2} {m\lambda\alpha}}  + \Big({\frac {(4m-2) d \tau^2 \rho^2 v^2 n^2} {m\lambda\alpha}}  + \gamma \Big) \sqrt{\frac {\ln {\delta^{-1}}} {2m}}$ in (\ref{eq23}) as ``overfit-bound", and reducing the overfit-bound is crucial for reducing the generalization risk bound. Most importantly, we can see from the overfit-bound that the structure regularization factor $\alpha$ is always staying together with the weight regularization factor $\lambda$, working together on reducing the overfit-bound. This indicates that the structure regularization is as important as the weight regularization for reducing the generalization risk for structured prediction.

Moreover, since $\tau, \rho$, and $v$ are typically small compared with other variables, especially $m$, (\ref{eq23}) can be approximated as follows by ignoring small terms:
\begin{equation}\label{eq24}
R(f) \leq R_e(f) + O\Big(\frac {d n^2 \sqrt {\ln {\delta^{-1}}}} {\lambda\alpha \sqrt m}  \Big)
\end{equation}
First, (\ref{eq24}) suggests that structure complexity $n$ can increase the overfit-bound on a magnitude of $O(n^2)$, and applying weight regularization can reduce the overfit-bound by $O(\lambda)$. Importantly, applying structure regularization further (over weight regularization) can additionally reduce the overfit-bound by a magnitude of $O(\alpha)$. When $\alpha=1$, it means ``no structure regularization'', then we have the worst overfit-bound.
Also, (\ref{eq24}) suggests that increasing the size of training set can reduce the overfit-bound on a square root level.

Theorem \ref{theo3} also indicates that too simple structures may overkill the overfit-bound but with a dominating empirical risk, and too complex structures may overkill the empirical risk but with a dominating overfit-bound. Thus, to achieve the best prediction accuracy, a balanced complexity of structures should be used for training the model.

\subsection{Accelerating Convergence Rates in Training}\label{sec:converge}

We also analyze the impact on the convergence rate of online learning by applying structure regularization. Our analysis is based on the stochastic gradient descent (SGD) setting \cite{bottou-98x,conf/nips/ZinkevichSL09,conf/nips/RechtRWN11}, which is arguably the most representative online training setting.
Let $g(\pmb w)$ be the structured prediction objective function and $\pmb w \in \mathcal W$ is the weight vector. Recall that the SGD update with fixed learning rate $\eta$ has a form like this:
\begin{equation}\label{eq29}
\pmb w_{t+1} \leftarrow \pmb w_t - \eta \nabla g_{\pmb z_t}(\pmb w_t)
\end{equation}
where $g_{\pmb z}(\pmb w_t)$ is the stochastic estimation of the objective function based on $\pmb z$ which is randomly drawn from $S$.

To state our convergence rate analysis results, we need several assumptions following (Nemirovski et al. 2009). We assume $g$ is strongly convex with modulus $c$, that is, $\forall \pmb w, \pmb w' \in \mathcal W$,
\begin{equation}\label{eq30}
g(\pmb w')\geq g(\pmb w)+ (\pmb w' -\pmb w)^T \nabla g(\pmb w) + \frac c 2 ||\pmb w' - \pmb w ||^2
\end{equation}
When $g$ is strongly convex, there is a global optimum/minimizer $\pmb w^*$.
We also assume Lipschitz continuous differentiability of $g$ with the constant $q$, that is, $\forall \pmb w, \pmb w' \in \mathcal W$,
\begin{equation}\label{eq31}
||\nabla g(\pmb w') - \nabla g(\pmb w)|| \leq q||\pmb w' - \pmb w||
\end{equation}
It is also reasonable to assume that the norm of $\nabla g_{\pmb z}(\pmb w)$ has almost surely positive correlation with the structure complexity of $\pmb z$,\footnote{Many structured prediction systems (e.g., CRFs) satisfy this assumption that the gradient based on a larger sample (i.e., $n$ is large) is expected to have a larger norm.} which can be quantified by a bound $\kappa \in \mathbb R^+$:
\begin{equation}\label{eq32}
||\nabla g_{\pmb z}(\pmb w) ||_2 \leq \kappa |\pmb z| \ \ \ \text{almost surely for} \ \ \ \forall \pmb w \in \mathcal W
\end{equation}
where $|\pmb z|$ denotes the structure complexity of $\pmb z$.
Moreover, it is reasonable to assume
\begin{equation}\label{eq33}
\eta c < 1
\end{equation}
because even the ordinary gradient descent methods will diverge if $\eta c >1$.

Then, we show that structure regularization can quadratically accelerate the SGD rates of convergence:
%
\begin{proposition}[Convergence rates vs. structure regularization]\label{theo4}
With the aforementioned assumptions, let the SGD training have a learning rate defined as
$
\eta = \frac {c\epsilon \beta \alpha^2} {q \kappa^2 n^2}
$,
where $\epsilon > 0$ is a convergence tolerance value and $\beta \in (0,1]$. Let $t$ be a integer satisfying
\begin{equation}\label{eq43}
t \geq \frac {q\kappa^2 n^2\log{(q a_0/\epsilon)}} {\epsilon \beta c^2 \alpha^2}
\end{equation}
where $n$ and $\alpha \in [1,n]$ is like before, and $a_0$ is the initial distance which depends on the initialization of the weights $\pmb w_0$ and the minimizer $\pmb w^*$, i.e., $a_0=||\pmb w_0 - \pmb w^*||^2$.
Then, after $t$ updates of $\pmb w$ it converges to $\mathbb E[g(\pmb w_t)-g(\pmb w^*)]\leq \epsilon$.
\end{proposition}

The proof is in Section \ref{proof}.

This Proposition demonstrates the $1/t$ convergence rate with $t$ given in (\ref{eq43}).
Recall that when $\alpha=1$, the algorithm with structure regularization reduces exactly to the ordinary algorithm (without structure regularization), which has the number of SGD updates $t \geq
\frac {q \kappa^2 n^2 \log {(q a_0 / \epsilon)}} {\epsilon \beta c^2} $ to achieve the convergence tolerance value $\epsilon$. In other words, applying structure regularization with the strength $\alpha$ can quadratically accelerate the convergence rate with a factor of $\alpha^2$.

\section{Experiments}

\subsection{Tasks}
\textbf{Diversified Tasks.} We experiment on natural language processing tasks and signal processing tasks. The natural language processing tasks include (1) part-of-speech tagging, (2) biomedical named entity recognition, and (3) Chinese word segmentation. The signal processing task is (4) sensor-based human activity recognition.
The tasks (1) to (3) use boolean features and the task (4) adopts real-valued features.
From tasks (1) to (4), the averaged structure complexity (number of observations) $n$ is very different, with $n=23.9, 26.5, 46.6, 67.9$, respectively. The dimension of tags $|\mathcal Y|$ is also diversified among tasks, with $|\mathcal Y|$ ranging from 5 to 45.

\textbf{Part-of-Speech Tagging (POS-Tagging).}
Part-of-Speech (POS) tagging is an important and highly competitive task in natural language processing. We use the standard benchmark dataset in prior work \cite{Collins2002}, which is derived from PennTreeBank corpus and uses sections 0 to 18 of the Wall Street Journal (WSJ) for training (38,219 samples), and sections 22-24 for testing (5,462 samples).
Following prior work \cite{Tsuruoka2011}, we use features based on unigrams and bigrams of neighboring words, and lexical patterns of the current word, with 393,741 raw features\footnote{Raw features are those observation features based only on $\pmb x$, i.e., no combination with tag information.} in total. Following prior work, the evaluation metric for this task is per-word accuracy.

\textbf{Biomedical Named Entity Recognition (Bio-NER).}
This task is from the \emph{BioNLP-2004 shared task}, which is for
recognizing 5 kinds of biomedical named entities (\emph{DNA}, \emph{RNA}, etc.) on the \emph{MEDLINE} biomedical text corpus. There are 17,484 training samples and 3,856 test samples.
Following prior work \cite{Tsuruoka2011}, we use word pattern features and POS features, with 403,192 raw features in total.
The evaluation metric is balanced F-score.

\textbf{Word Segmentation (Word-Seg).}
Chinese word segmentation is important and it is usually the first step for text processing in Chinese.
We use the Microsoft Research (MSR) data provided by \emph{SIGHAN-2004 contest}.
There are 86,918 training samples and 3,985 test samples.
Following prior work \cite{GaoAndrew07}, we use features based on character unigrams and bigrams, with 1,985,720 raw features in total.
The evaluation metric for this task is balanced F-score.

\textbf{Sensor-based Human Activity Recognition (Act-Recog).}
This is a task based on real-valued sensor signals, with the data extracted from the Bao04 activity recognition dataset~\cite{tkde/SunKU13}.
This task aims to recognize human activities (walking, bicycling, etc.) by using 5 biaxial sensors to collect acceleration signals of individuals, with the sampling frequency at 76.25HZ.
Following prior work in activity recognition~\cite{tkde/SunKU13}, we use acceleration features, mean features, standard deviation, energy, and correlation features, with 1228 raw features in total.
There are 16,000 training samples and 4,000 test samples.
Following prior work, the evaluation metric is accuracy.

\subsection{Experimental Settings}

To test the robustness of the proposed structure regularization (\emph{StructReg}) method, we perform experiments on both probabilistic and non-probabilistic structure prediction models. We choose the conditional random fields (CRFs) \cite{LaffertyMcCallum01} and structured perceptrons (Perc) \cite{Collins2002}, which are arguably the most popular probabilistic and non-probabilistic structured prediction models, respectively. The CRFs are trained using the SGD algorithm,\footnote{In theoretical analysis, following prior work we adopt the SGD with fixed learning rate, as described in Section \ref{sec:converge}. However, since the SGD with decaying learning rate is more commonly used in practice, in experiments we use the SGD with decaying learning rate.} and the baseline method is the traditional weight regularization scheme (\emph{WeightReg}), which adopts the most representative $L_2$ weight regularization, i.e.,  a Gaussian prior.\footnote{We also tested on sparsity emphasized regularization methods, including $L_1$ regularization and \emph{Group Lasso} regularization \cite{MartinsSFA11a}. However, we find that in most cases those sparsity emphasized regularization methods have lower accuracy than the $L_2$ regularization.} For the structured perceptrons, the baseline \emph{WeightAvg} is the popular implicit regularization technique based on parameter averaging, i.e., \emph{averaged perceptron} \cite{Collins2002}.

All methods use the same set of features. Since the rich edge features \cite{conf/acl/SunWL12,SunLWL14} can be automatically generated from raw features and are very useful for improving model accuracy, the rich edge features are employed for all methods. All methods are based on the 1st-order Markov dependency.
For \emph{WeightReg}, the $L_2$ regularization strengths (i.e., $\lambda/2$ in Eq.\ref{eq22}) are tuned among values $0.1, 0.5, 1, 2, 5$, and are determined on the development data provided by the standard dataset (POS-Tagging) or simply via 4-fold cross validation on the training set (Bio-NER, Word-Seg, and Act-Recog). With this automatic tuning for \emph{WeightReg}, we set $2, 5, 1$ and $5$ for POS-Tagging, Bio-NER, Word-Seg, and Act-Recog tasks, respectively. Our \emph{StructReg} method adopts the same $L_2$ regularization setting like \emph{WeightReg}.
Experiments are performed on a computer with Intel(R) Xeon(R) 3.0GHz CPU.

\subsection{Experimental Results}

\begin{figure*}[tb]
\centering
\begin{tabular}{@{}c@{}@{}c@{}@{}c@{}@{}c@{}}

\epsfig{file=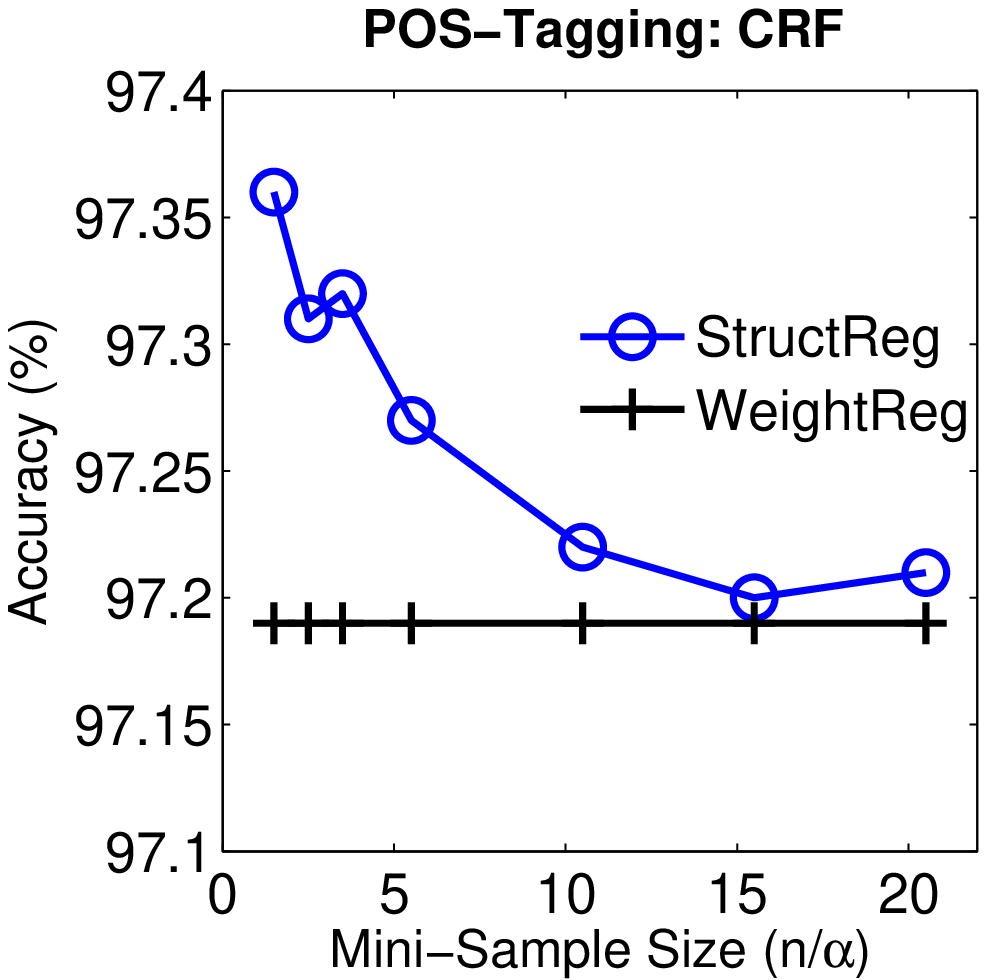,width=0.25\linewidth,clip=} &
\epsfig{file=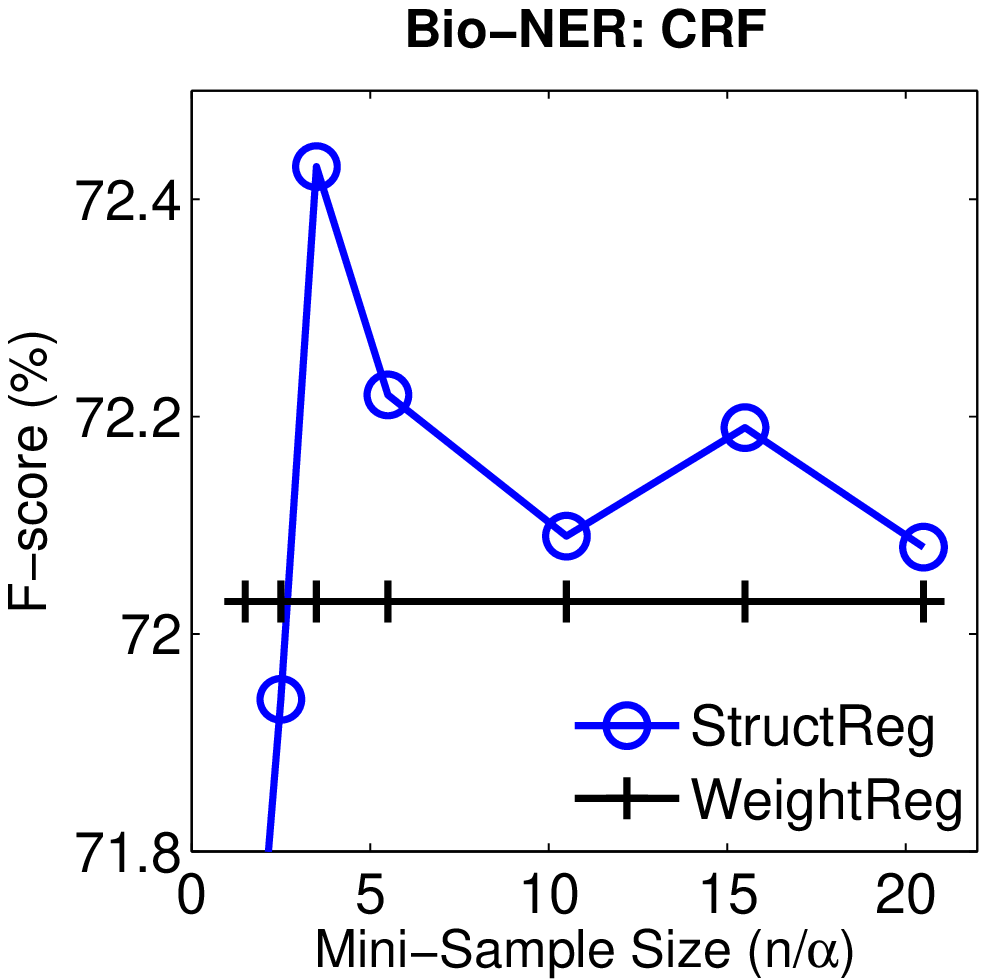,width=0.25\linewidth,clip=} &
\epsfig{file=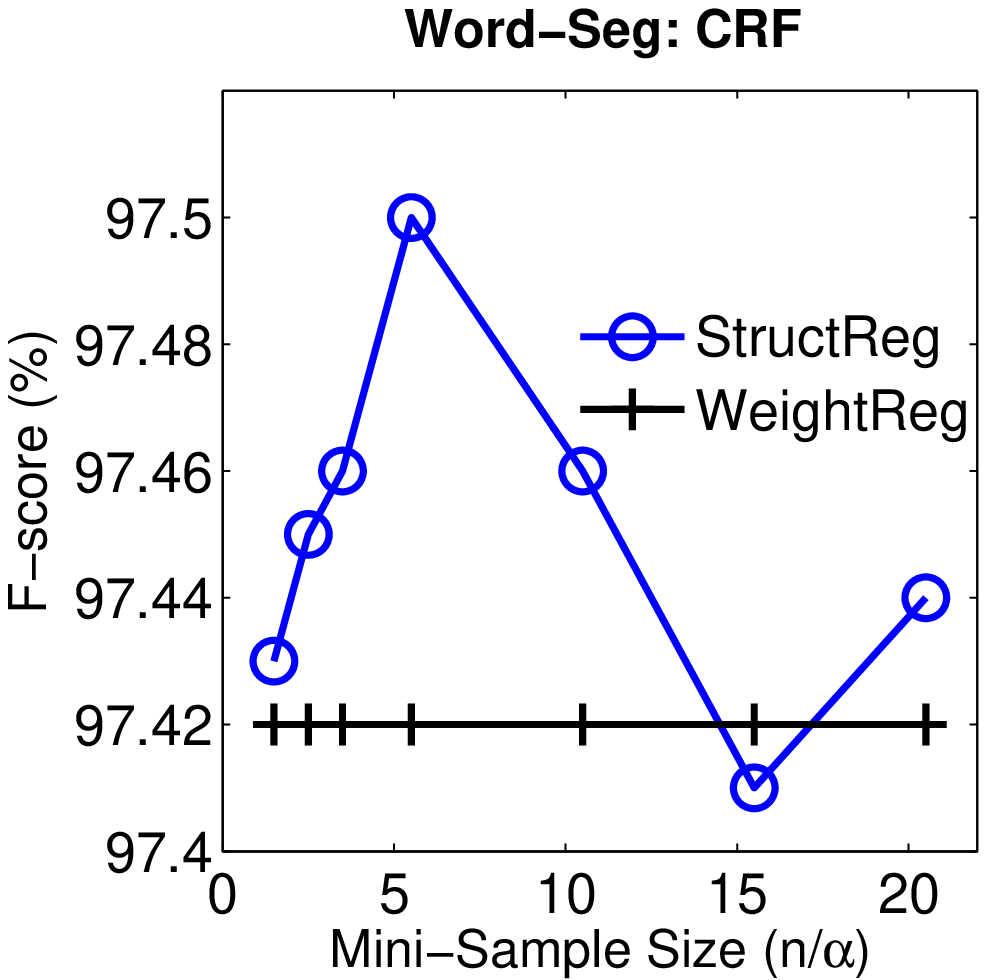,width=0.25\linewidth,clip=} &
\epsfig{file=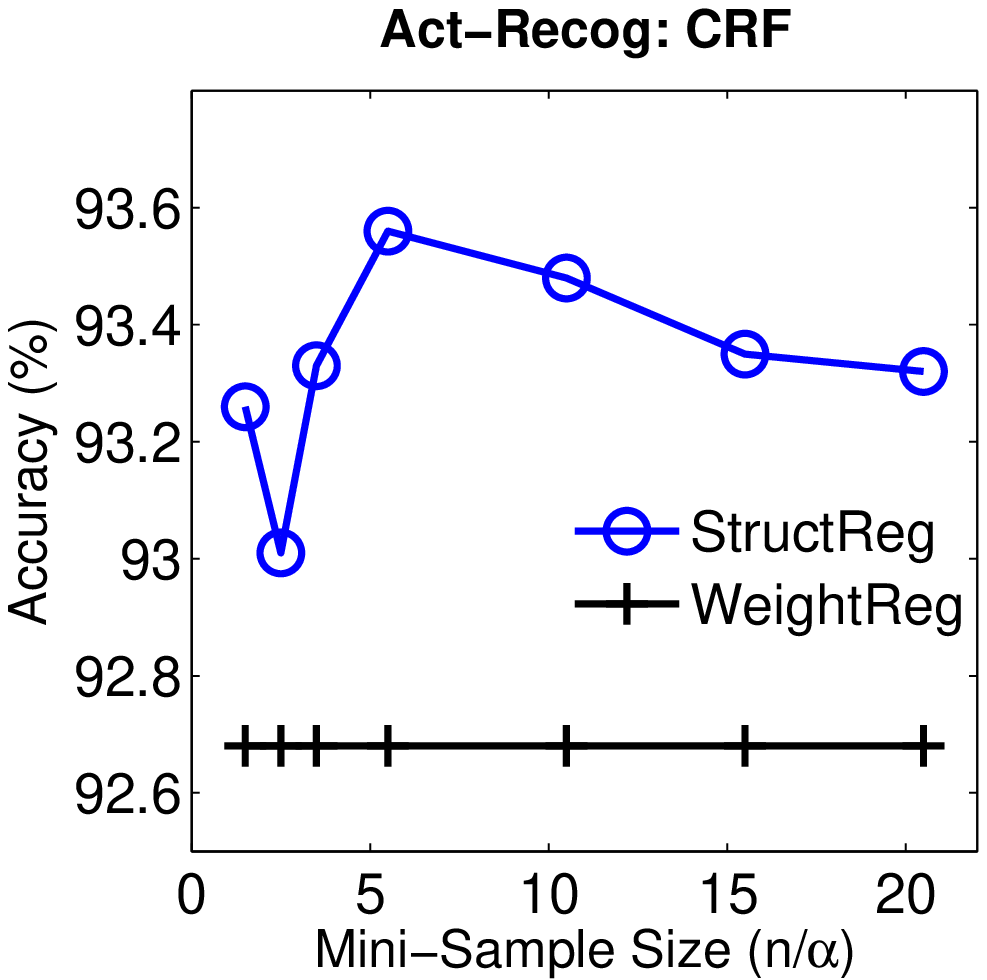,width=0.25\linewidth,clip=} \\

\epsfig{file=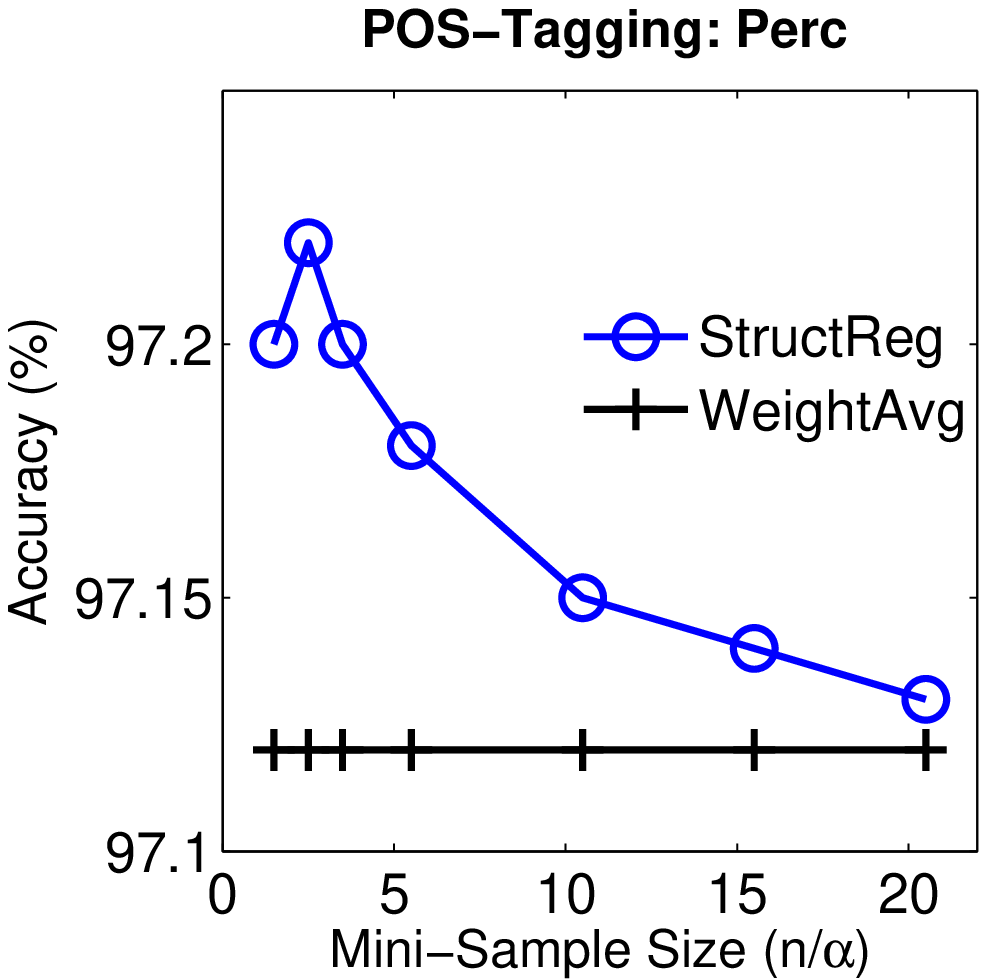,width=0.25\linewidth,clip=} &
\epsfig{file=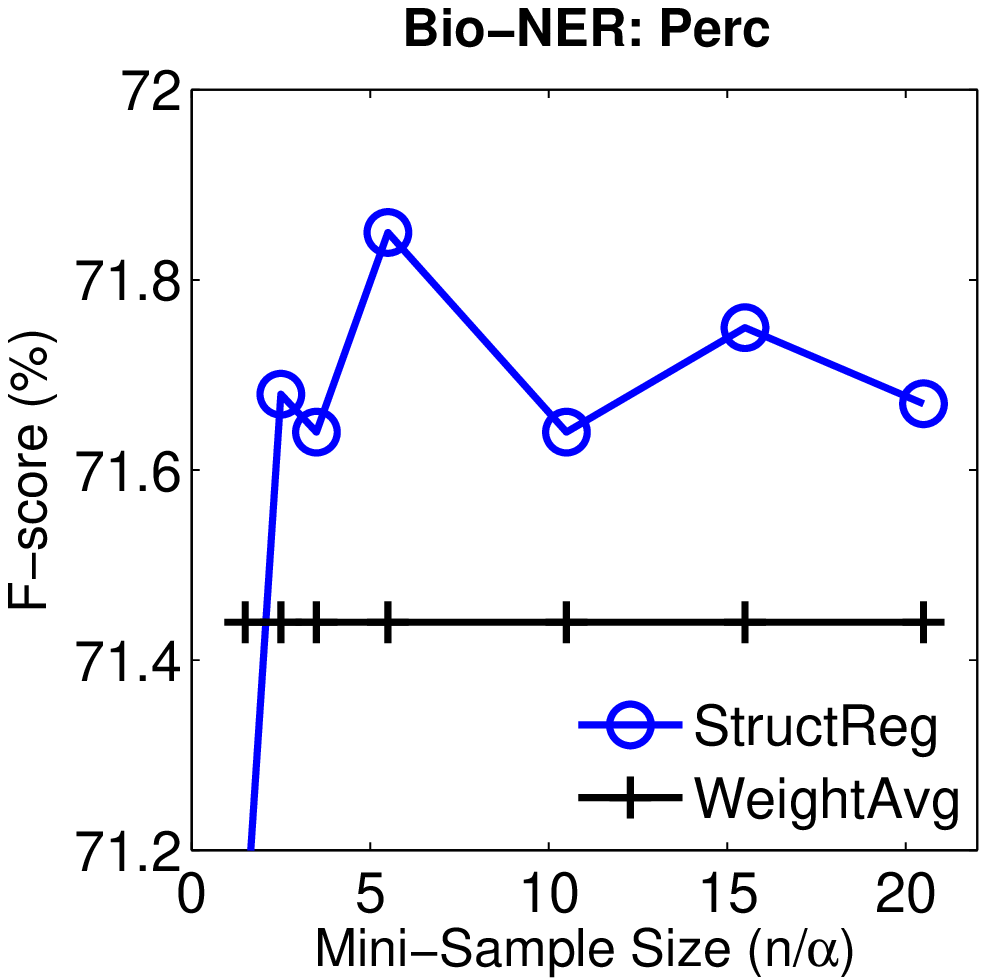,width=0.25\linewidth,clip=} &
\epsfig{file=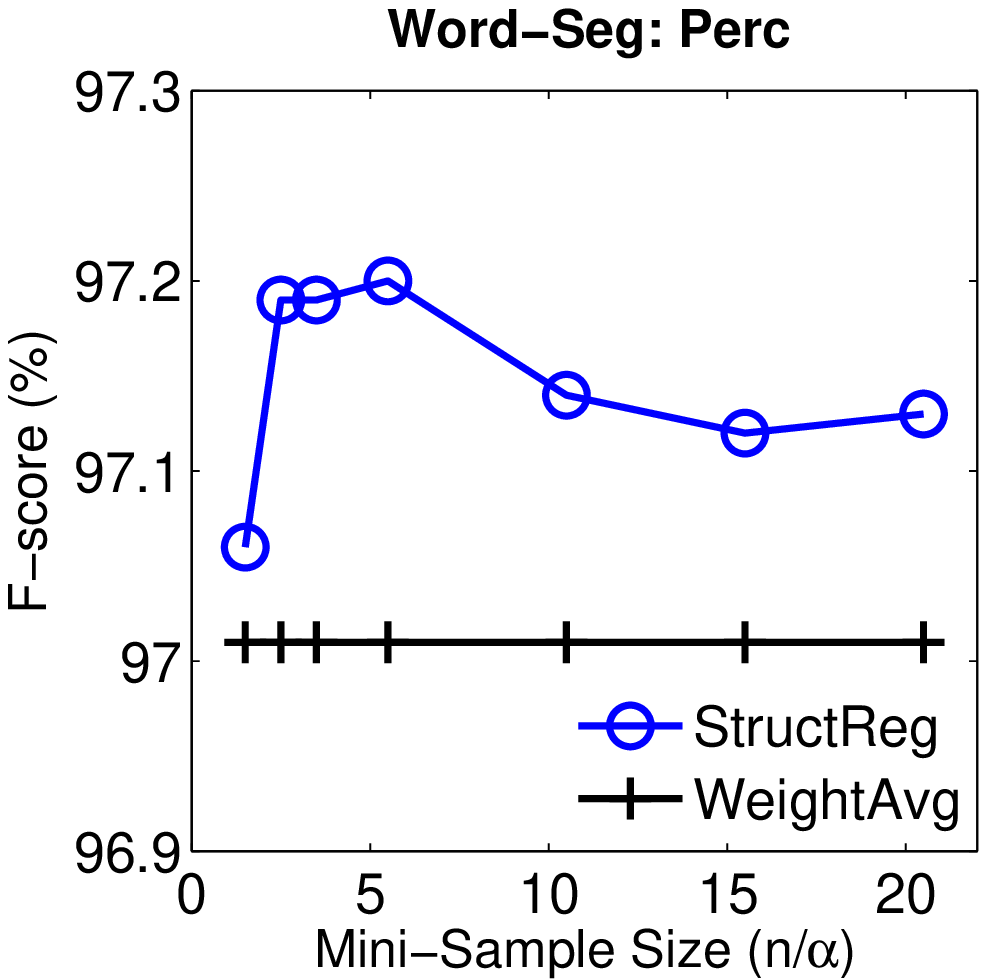,width=0.25\linewidth,clip=} &
\epsfig{file=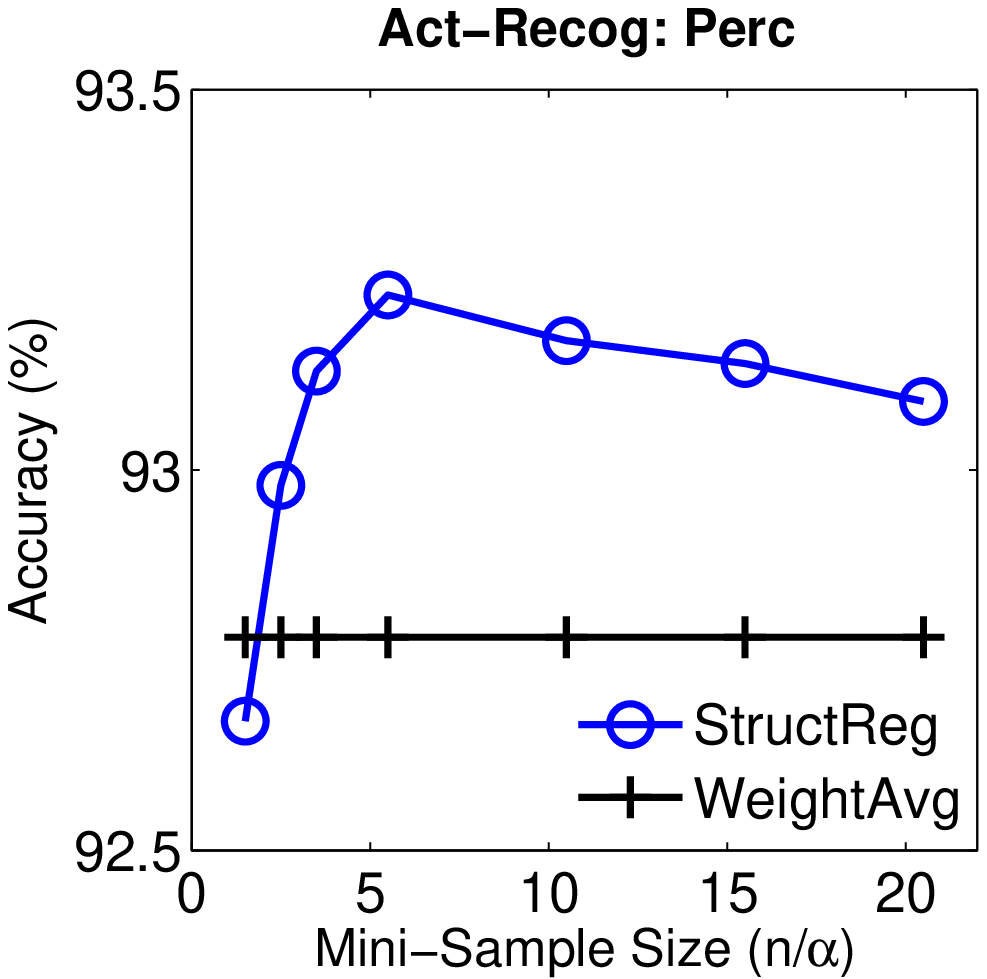,width=0.25\linewidth,clip=} \\

\end{tabular}
\caption{On the four tasks, comparing the structure regularization method (\emph{StructReg}) with existing regularization methods in terms of accuracy/F-score. Row-1 shows the results on CRFs and Row-2 shows the results on structured perceptrons.
}\label{fig.acc}
\vspace{-0.1in}
\end{figure*}

\begin{table}[t]
\centering
\caption{Comparing our results with the benchmark systems on corresponding tasks.}\label{tab:art}
\begin{tabular}{ccccc} \hline
&POS-Tagging (Acc\%) &Bio-NER (F1\%) &Word-Seg (F1\%)       \\ \hline
Benchmark system &97.33 (see \cite{acl/ShenSJ07}) &72.28 (see \cite{Tsuruoka2011}) &97.19 (see \cite{GaoAndrew07})  \\
Our results &\textbf{97.36} &\textbf{72.43} &\textbf{97.50} \\ \hline
\end{tabular}
\vspace{-0.15in}
\end{table}

\begin{figure*}[tb]
\centering
\begin{tabular}{@{}c@{}@{}c@{}@{}c@{}@{}c@{}}

\epsfig{file=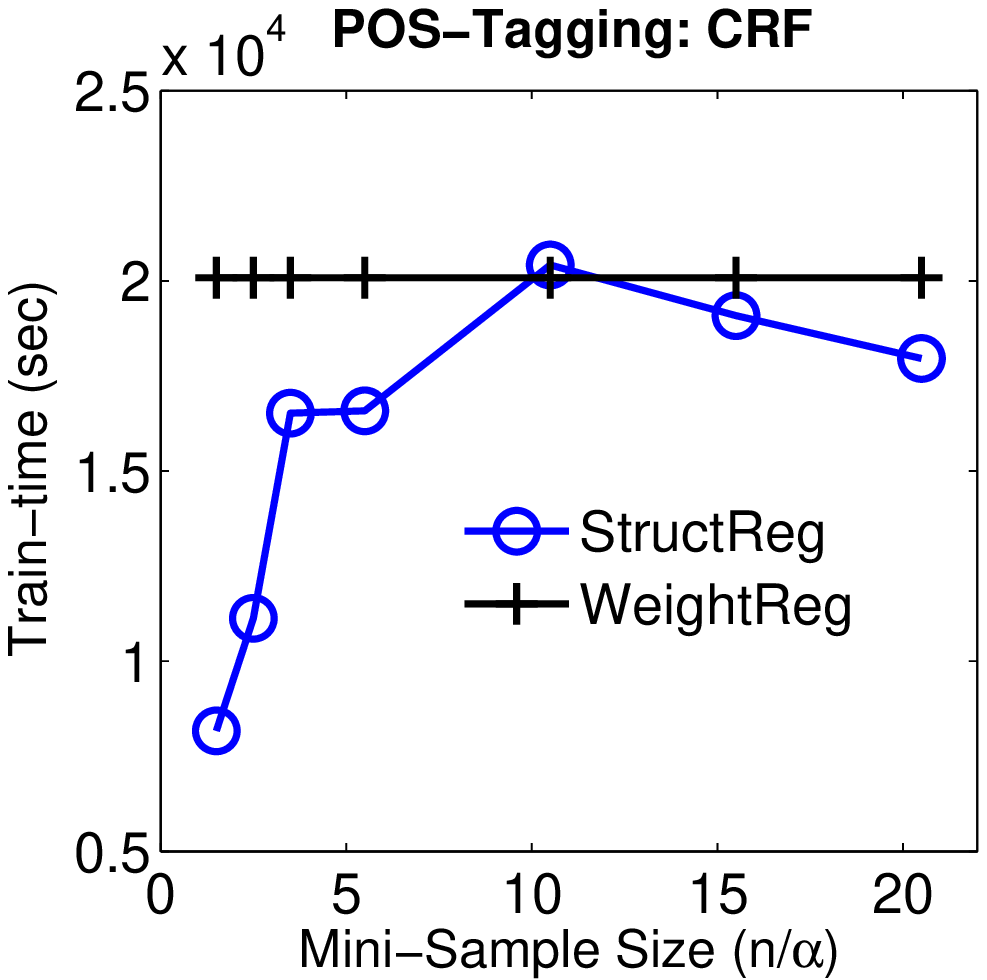,width=0.25\linewidth,clip=} &
\epsfig{file=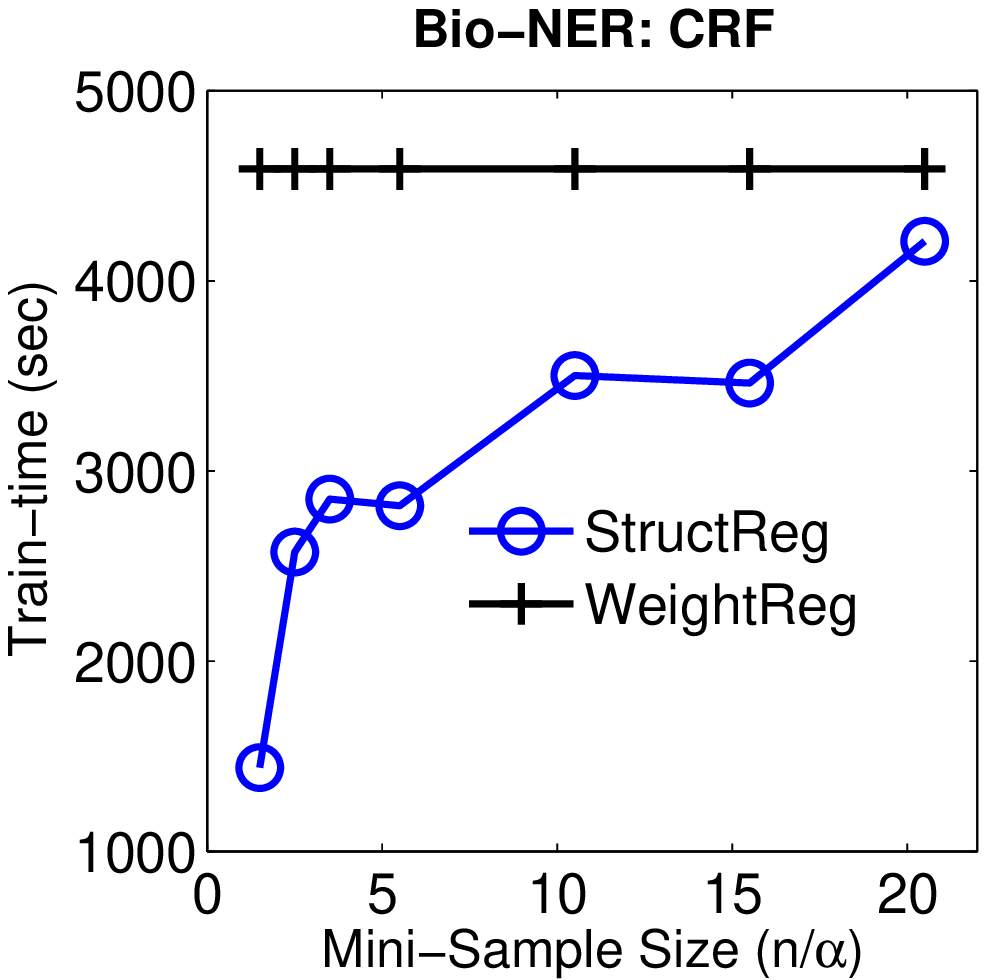,width=0.25\linewidth,clip=} &
\epsfig{file=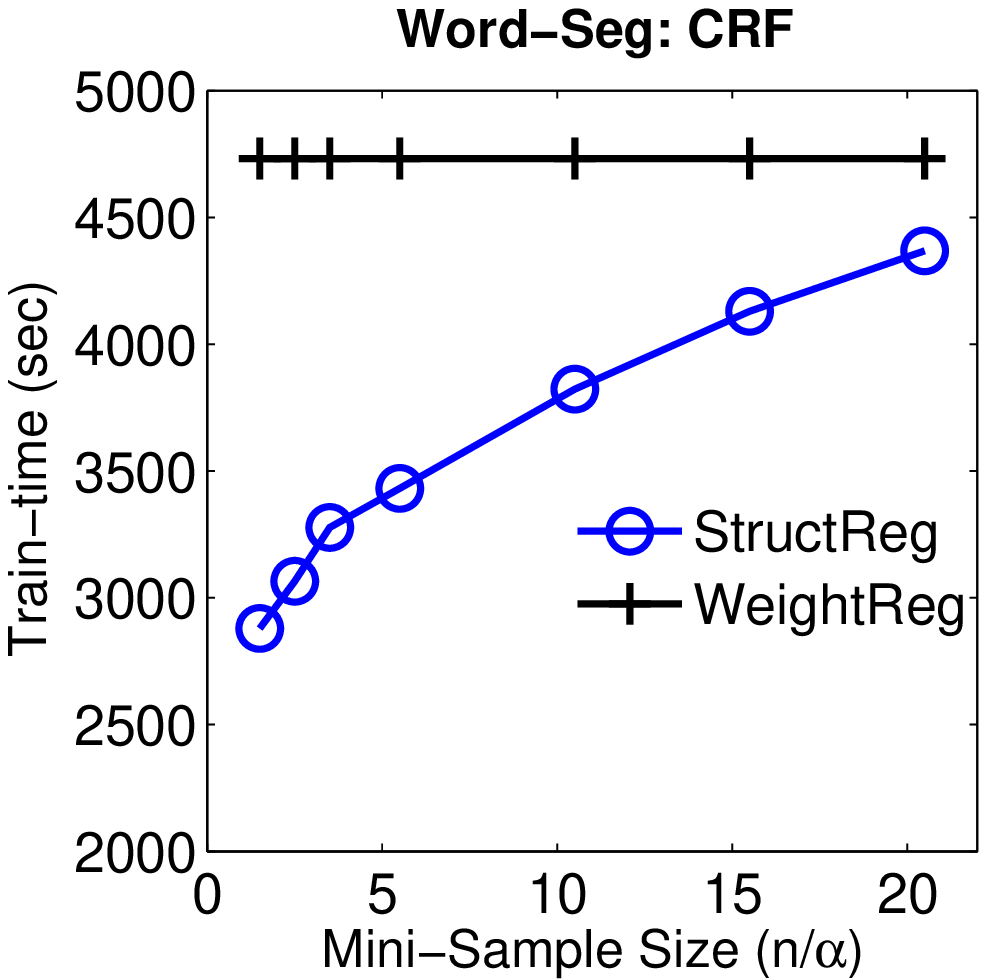,width=0.25\linewidth,clip=} &
\epsfig{file=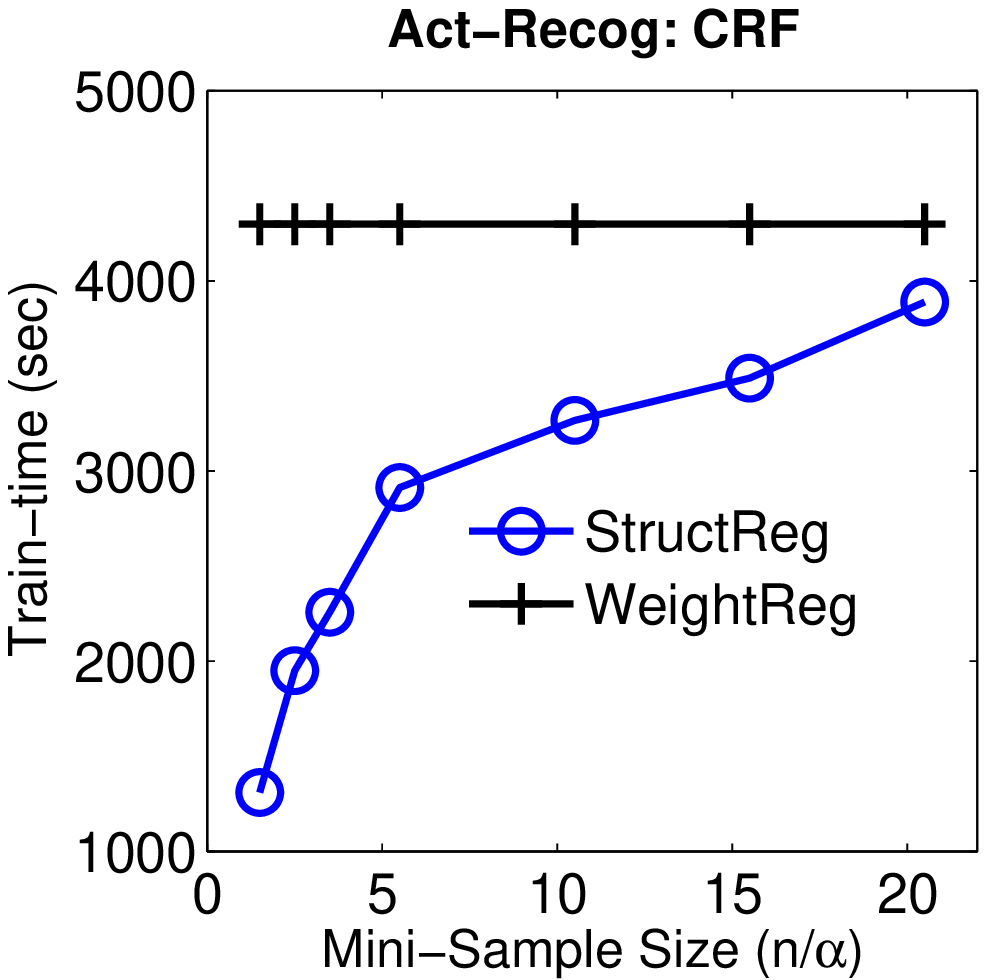,width=0.25\linewidth,clip=} \\

\epsfig{file=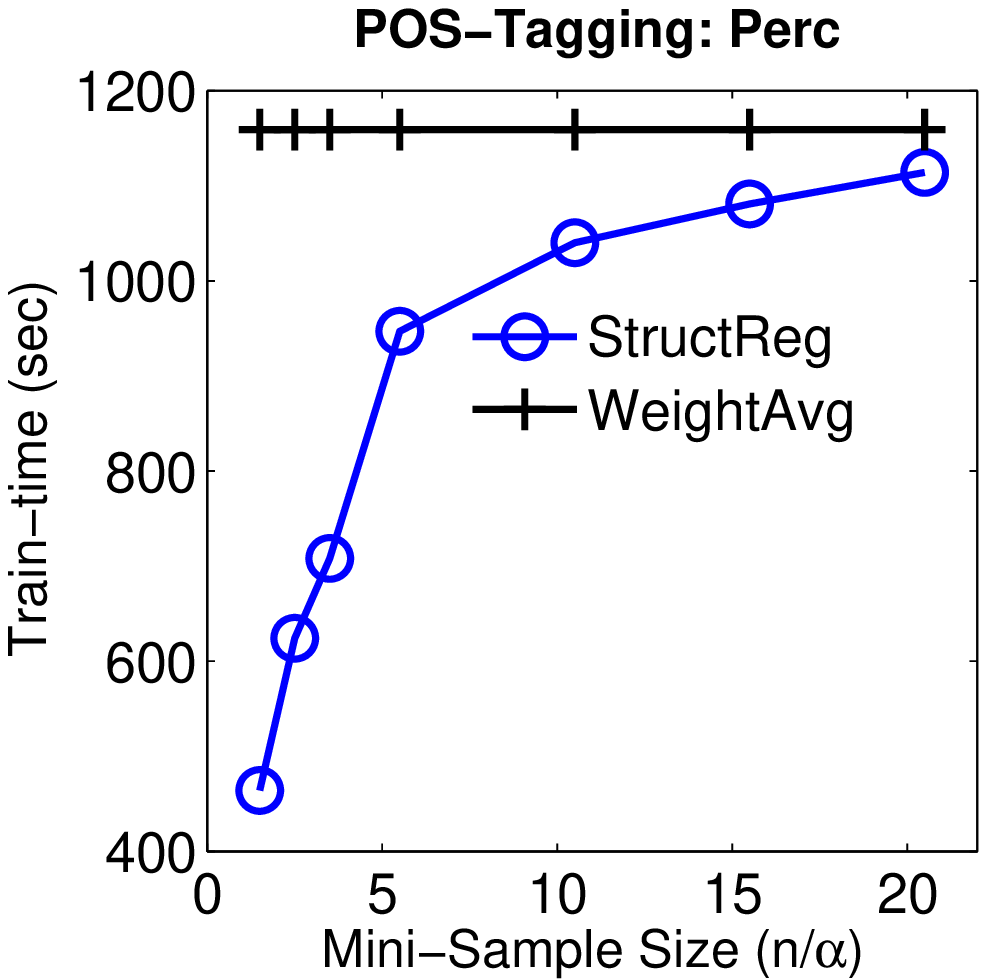,width=0.25\linewidth,clip=} &
\epsfig{file=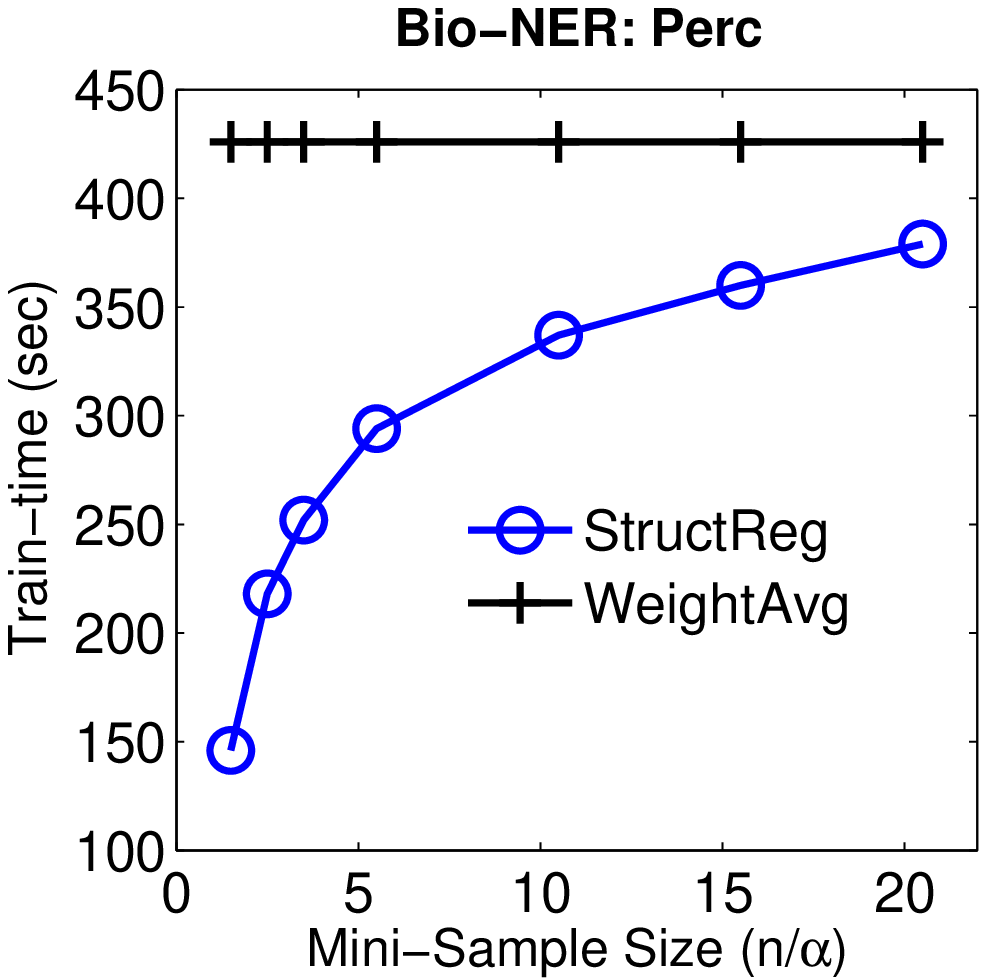,width=0.25\linewidth,clip=} &
\epsfig{file=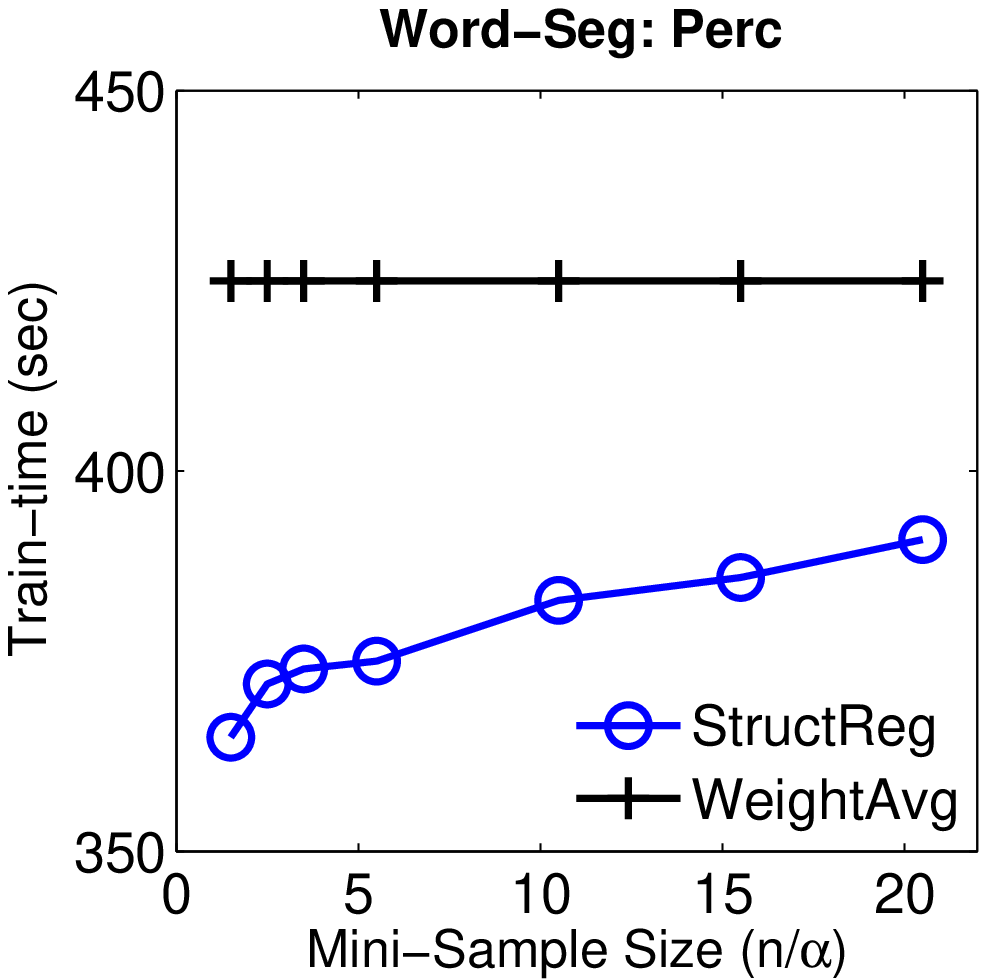,width=0.25\linewidth,clip=} &
\epsfig{file=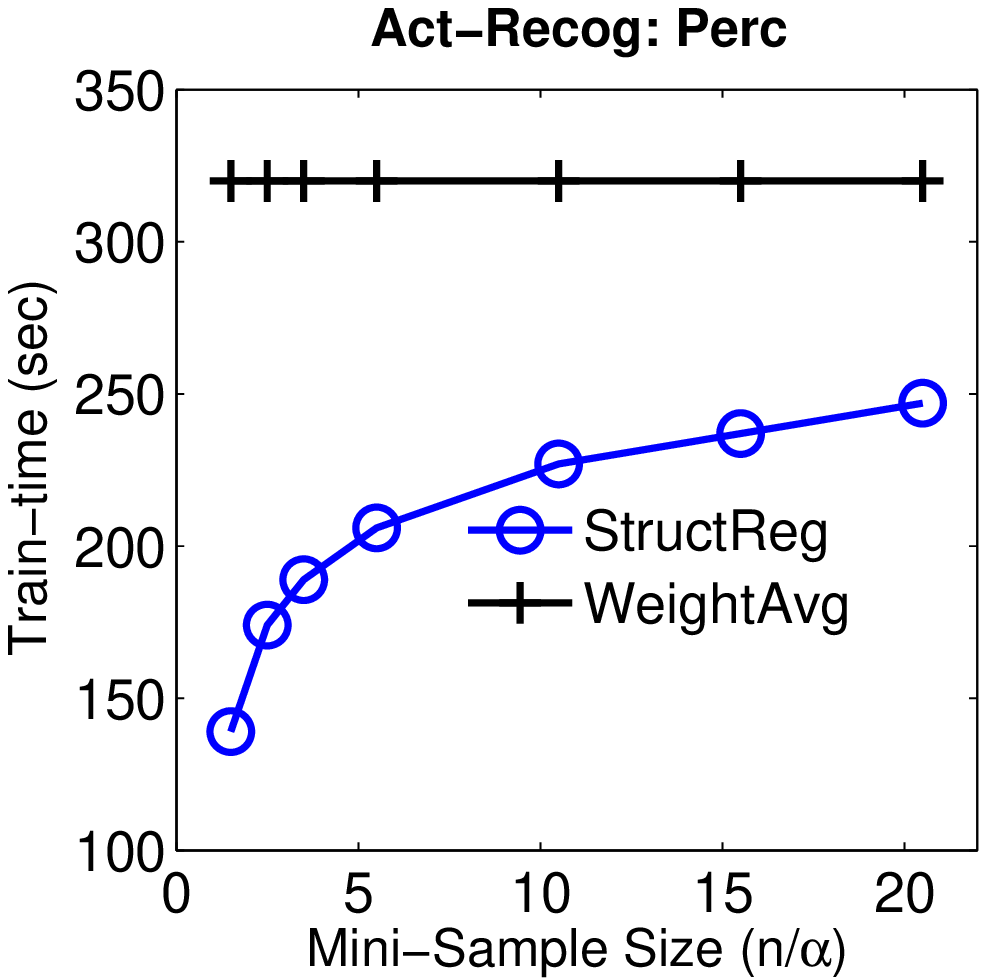,width=0.25\linewidth,clip=} \\

\end{tabular}
\caption{On the four tasks, comparing the structure regularization method (\emph{StructReg}) with existing regularization methods in terms of wall-clock training time.
}\label{fig.time}
\vspace{-0.15in}
\end{figure*}

The experimental results in terms of accuracy/F-score are shown in Figure \ref{fig.acc}. For the CRF model, the training is convergent, and the results on the convergence state (decided by relative objective change with the threshold value of $0.0001$) are shown.
For the structured perceptron model, the training is typically not convergent, and the results on the 10'th iteration are shown. For stability of the curves, the results of the structured perceptrons are averaged over 10 repeated runs.

Since different samples have different size $n$ in practice, we set $\alpha$ being a function of $n$, so that the generated mini-samples are with \emph{fixed} size $n'$ with $n'=n/\alpha$. Actually, $n'$ is a probabilistic distribution because we adopt randomized decomposition. For example, if $n'=5.5$, it means the mini-samples are a mixture of the ones with the size 5 and the ones with the size 6, and the mean of the size distribution is 5.5.
In the figure, the curves are based on $n'=1.5, 2.5, 3.5, 5.5, 10.5, 15.5, 20.5$.

As we can see, although the experiments are based on very different models (probabilistic or non-probabilistic), with diversified feature types (boolean or real-value) and different structure complexity $n$, the results are quite consistent. It demonstrates that structure regularization leads to higher accuracies/F-scores compared with the existing baselines.

We also conduct significance tests based on t-test.
Since the t-test for F-score based tasks (Bio-NER and Word-Seg) may be unreliable\footnote{Indeed we can convert F-scores to accuracy scores for t-test, but in many cases this conversion is unreliable. For example, very different F-scores may correspond to similar accuracy scores.}, we only perform t-test for the accuracy-based tasks, i.e., POS-Tagging and Act-Recog. For POS-Tagging, the significance test suggests that the superiority of \emph{StructReg} over \emph{WeightReg} is very statistically significant, with $p<0.01$. For Act-Recog, the significance tests suggest that both the \emph{StructReg vs. WeightReg} difference and the \emph{StructReg vs. WeightAvg} difference are extremely statistically significant, with $p<0.0001$ in both cases.
The experimental results support our theoretical analysis that structure regularization can further reduce the generalization risk over existing weight regularization techniques.

Our method actually outperforms the benchmark systems on the three important natural language processing tasks.
The POS-Tagging task is a highly competitive task, with many methods proposed, and the best report (without using extra resources) until now is achieved by using a bidirectional learning model in \cite{acl/ShenSJ07},\footnote{See a collection of the systems at \url{http://aclweb.org/aclwiki/index.php?title=POS_Tagging_(State_of_the_art)}} with the accuracy 97.33\%. Our simple method achieves better accuracy compared with all of those state-of-the-art systems. Furthermore, our method achieves as good scores as the benchmark systems on the Bio-NER and Word-Seg tasks, which are also very competitive tasks in natural language processing communities. On the Bio-NER task, \cite{Tsuruoka2011} achieves 72.28\% based on lookahead learning and \cite{Yoshida07} achieves 72.65\% based on reranking. On the Word-Seg task, \cite{GaoAndrew07} achieves 97.19\% based on maximum entropy classification and our recent work \cite{SunLWL14} achieves 97.5\% based on feature-frequency-adaptive online learning. The comparisons are summarized in Table \ref{tab:art}. Note that, similar to the tuning on the \emph{WeightReg} strengths, the optimal values of \emph{StructReg} strengths are also decided automatically based on standard development data or cross validation on training data.

Figure \ref{fig.time} shows experimental comparisons in terms of wall-clock training time. As we can see, the proposed method can substantially improve the training speed. The speedup is not only from the faster convergence rates, but also from the faster processing time on the structures, because it is more efficient to process the decomposed samples with simple structures.

\section{Proofs}\label{proof}

Our analysis sometimes need to use McDiarmid's inequality.

\begin{theorem}[McDiarmid, 1989]\label{theo1}
Let $S= \{q_1, \dots, q_m\}$ be independent random variables taking values in the space $Q^m$. Moreover, let $g: Q^m \mapsto \mathbb R$ be a function of $S$ that satisfies $\forall i, \forall S \in Q^m, \forall \hat{q}_i \in Q, $
$$
| g(S) - g(S^i) | \leq c_i.
$$
Then $\forall \epsilon > 0$,
$$
\mathbb P_S[g(S)- \mathbb E_S[g(S)] \geq \epsilon] \leq \exp \Big (\frac {-2\epsilon^2} {\sum_{i=1}^{m} c_i^2} \Big ) .
$$
\end{theorem}

\begin{lemma}[Symmetric learning]\label{lemma1}
For any symmetric (i.e., order-free) learning algorithm $G$, $\forall i \in \{1, \dots, m\}$, we have
\begin{equation*}
\mathbb{E}_S [R(G_S) - R_{e}(G_S)]= \frac 1 n \mathbb{E}_{S, \pmb {\hat z}_i} [\mathcal L (G_S, \pmb {\hat z}_i) -  \mathcal L (G_{S^i}, \pmb {\hat z}_i)]
\end{equation*}
 \end{lemma}

\proof
\begin{equation*}
\begin{split}
\mathbb E_S [R(G_S) - R_{e}(G_S)]
&= \frac 1 n \mathbb E_S \Big (\mathbb E_{\pmb z}(\mathcal L (G_S, \pmb z))- \frac 1 m \sum_{j=1}^m \mathcal L(G_S, \pmb z_j) \Big ) \\
&= \frac 1 n \Big( \mathbb E_{S, \pmb {\hat z}_i} \big( \mathcal L(G_S, \pmb {\hat z}_i) \big) - \frac 1 m \sum_{j=1}^m \mathbb E_S \big( \mathcal L(G_S, \pmb z_j) \big) \Big)\\
&= \frac 1 n \Big( \mathbb E_{S, \pmb {\hat z}_i} \big( \mathcal L(G_S, \pmb {\hat z}_i) \big) - \mathbb E_S \big( \mathcal L(G_S, \pmb z_i) \big) \Big)\\
&= \frac 1 n \Big( \mathbb E_{S, \pmb {\hat z}_i} \big( \mathcal L(G_S, \pmb {\hat z}_i) \big) - \mathbb E_{S^i} \big( \mathcal L(G_{S^i}, \pmb {\hat z}_i) \big) \Big)\\
&= \frac 1 n \mathbb{E}_{S, \pmb {\hat z}_i} \big( \mathcal L (G_S, \pmb {\hat z}_i) -  \mathcal L (G_{S^i}, \pmb {\hat z}_i) \big)
\end{split}
\end{equation*}
where the 3rd step is based on $\mathbb E_S \mathcal L(G_S, \pmb z_i)= \mathbb E_S \mathcal L(G_S, \pmb z_j)$ for $\forall \pmb z_i \in S$ and $\forall \pmb z_j \in S$, given that $G$ is symmetric.
\endproof

\subsection{Proofs}

\textbf{Proof of Lemma \ref{lemma2}}

According to (\ref{eq12}), we have $\forall i, \forall S, \forall \pmb z, \forall k$
\begin{equation*}
\begin{split}
|\ell_\tau(G_S,\pmb z, k) - \ell_\tau (G_{S^{\setminus i}}, \pmb z, k)|
&= |c_\tau [G_S(\pmb x, k), \pmb y_{(k)}] - c_\tau [G_{S^{\setminus i}} (\pmb x, k), \pmb y_{(k)}]|\\
&\leq \tau | G_S(\pmb x, k) - G_{S^{\setminus i}} (\pmb x, k)|\\
&\leq \tau \Delta
\end{split}
\end{equation*}
This gives the bound of loss stability.

Also, we have $\forall i, \forall S, \forall \pmb z$
\begin{equation*}
\begin{split}
|\mathcal L_\tau(G_S,\pmb z) - \mathcal L_\tau (G_{S^{\setminus i}}, \pmb z)|
&= \Big |\sum_{k=1}^n c_\tau [G_S(\pmb x, k), \pmb y_{(k)}] - \sum_{k=1}^n c_\tau [G_{S^{\setminus i}} (\pmb x, k), \pmb y_{(k)}] \Big |\\
&\leq \sum_{k=1}^n \Big | c_\tau [G_S(\pmb x, k), \pmb y_{(k)}] - c_\tau [G_{S^{\setminus i}} (\pmb x, k), \pmb y_{(k)}] \Big |\\
&\leq \tau \sum_{k=1}^n  | G_S(\pmb x, k) - G_{S^{\setminus i}} (\pmb x, k)|\\
&\leq n\tau\Delta
\end{split}
\end{equation*}
This derives the bound of sample loss stability.
\endproof

\textbf{Proof of Theorem \ref{theo1.2}}

When a convex and differentiable function $g$ has a minimum $f$ in space $\mathcal F$, its Bregman divergence has the following property for $\forall f' \in \mathcal F$:
$$
d_g(f',f)=g(f')-g(f)
$$
With this property, we have
\begin{equation}\label{eq13}
\begin{split}
d_{R_{\alpha,\lambda}}(f^{\setminus i'},f)+d_{R_{\alpha,\lambda}^{\setminus i'}}(f,f^{\setminus i'}) &= R_{\alpha,\lambda}(f^{\setminus i'}) - R_{\alpha,\lambda}(f) +R_{\alpha,\lambda}^{\setminus i'}(f) - R_{\alpha,\lambda}^{\setminus i'}(f^{\setminus i'})\\
&= \big( R_{\alpha,\lambda}(f^{\setminus i'}) - R_{\alpha,\lambda}^{\setminus i'}(f^{\setminus i'}) \big) - \big( R_{\alpha,\lambda}(f) -R_{\alpha,\lambda}^{\setminus i'}(f) \big)  \\
&= \frac 1 {mn} \mathcal L_\tau(f^{\setminus i'},\pmb z_{i'}') - \frac 1 {mn} \mathcal L_\tau(f,\pmb z_{i'}')
\end{split}
\end{equation}

Then, based on the property of Bregman divergence that $d_{g+g'}=d_g + d_{g'}$, we have
\begin{equation}\label{eq15}
\begin{split}
d_{N_\lambda}(f,f^{\setminus i'}) &+ d_{N_\lambda}(f^{\setminus i'},f)
= d_{(R_{\alpha,\lambda}^{\setminus i'} - R_\alpha^{\setminus i'})}(f,f^{\setminus i'}) + d_{(R_{\alpha,\lambda}-R_\alpha)}(f^{\setminus i'},f)\\
&=d_{R_{\alpha,\lambda}}(f^{\setminus i'},f) + d_{R_{\alpha,\lambda}^{\setminus i'}}(f,f^{\setminus i'}) - d_{R_\alpha}(f^{\setminus i'},f) - d_{R_\alpha^{\setminus i'}}(f,f^{\setminus i'})\\
&\text{(based on non-negativity of Bregman divergence)}\\
&\leq d_{R_{\alpha,\lambda}}(f^{\setminus i'},f) + d_{R_{\alpha,\lambda}^{\setminus i'}}(f,f^{\setminus i'})\\
&\text{(using (\ref{eq13}))}\\
&= \frac 1 {mn} \big( \mathcal L_\tau(f^{\setminus i'},\pmb z_{i'}') - \mathcal L_\tau(f,\pmb z_{i'}') \big)\\
&= \frac 1 {mn} \sum_{k=1}^{n/\alpha} \big( \ell_\tau(f^{\setminus i'},\pmb z_{i'}',k) - \ell_\tau(f,\pmb z_{i'}',k) \big)\\
&\leq \frac 1 {mn} \sum_{k=1}^{n/\alpha} \Big| c_\tau\Big(f^{\setminus i'}(\pmb x_{i'}',k), \pmb y_{i'(k)}'\Big) - c_\tau\Big(f(\pmb x_{i'}',k),\pmb y_{i'(k)}'\Big) \Big|\\
&\leq \frac \tau {mn} \sum_{k=1}^{n/\alpha}  \Big| f^{\setminus i'}(\pmb x_{i'}',k) - f(\pmb x_{i'}',k)  \Big|\\
&\text{(using (\ref{eq14}))}\\
&\leq \frac {\rho \tau } {m\alpha}   ||f-f^{\setminus i'}||_2 \cdot ||\pmb x_{i'}'||_2\\
\end{split}
\end{equation}

Moreover, $N_\lambda(g)=\frac \lambda {2} ||g||_2^2 =\frac \lambda {2} \langle g,g \rangle$ is a convex function and its Bregman divergence satisfies:
\begin{equation}\label{eq16}
\begin{split}
d_{N_\lambda}(g,g') &= \frac \lambda {2} \big(\langle g,g \rangle - \langle g',g' \rangle - \langle 2g',g-g' \rangle \big) \\
&= \frac \lambda {2} ||g-g'||_2^2
\end{split}
\end{equation}

Combining (\ref{eq15}) and (\ref{eq16}) gives
\begin{equation}\label{eq17}
\lambda ||f-f^{\setminus i'}||_2^2 \leq \frac {\rho \tau } {m\alpha}   ||f-f^{\setminus i'}||_2 \cdot ||\pmb x_{i'}'||_2
\end{equation}
which further gives
\begin{equation}\label{eq18}
||f-f^{\setminus i'}||_2 \leq \frac {\rho \tau } {m\lambda\alpha}   ||\pmb x_{i'}'||_2
\end{equation}

Given $\rho$-admissibility, we derive the bound of function stability $\Delta(f)$ based on sample $\pmb z$ with size $n$. We have  $\forall \pmb z=(\pmb x, \pmb y), \forall k,$
 \begin{equation}\label{eq19}
\begin{split}
|f(\pmb x, k) - f^{\setminus i'}(\pmb x, k)| &\leq  \rho||f- f^{\setminus i'}||_2\cdot ||\pmb x||_2\\
&\text{(using (\ref{eq18}))}\\
&\leq \frac {\tau \rho^2 } {m\lambda\alpha}   ||\pmb x_{i'}'||_2 \cdot ||\pmb x||_2
\end{split}
\end{equation}

With the feature dimension $d$ and $\pmb x_{(k,q)} \leq v$ for $q\in \{1,\dots,d\}$ , we have
 \begin{equation}\label{eq20}
\begin{split}
||\pmb x||_2 &=||\sum_{k=1}^n \pmb x_{(k)}||_2\\
&\leq || \langle \underbrace {nv, \dots, nv}_{d} \rangle ||_2\\
&= \sqrt{dn^2v^2}\\
&=nv\sqrt d
\end{split}
\end{equation}
Similarly, we have
$||\pmb x_{i'}'||_2 \leq \frac {nv\sqrt d} \alpha$ because $\pmb x_{i'}'$ is with the size $n/\alpha$.

Inserting the bounds of $||\pmb x||_2$ and $||\pmb x_{i'}'||_2$ into (\ref{eq19}), it goes to
\begin{equation}\label{eq21}
|f(\pmb x, k) - f^{\setminus i'}(\pmb x, k)| \leq \frac {d \tau \rho^2 v^2 n^2} {m\lambda\alpha^2}
\end{equation}
which gives (\ref{eq11}). Further, using Lemma \ref{lemma2} derives the loss stability bound of $\frac {d \tau^2 \rho^2 v^2 n^2} {m\lambda\alpha^2}$, and the sample loss stability bound of $\frac {d \tau^2 \rho^2 v^2 n^3} {m\lambda\alpha^2}$ on the minimizer $f$.
\endproof

\textbf{Proof of Corollary \ref{coro1}}

The proof is similar to the proof of Theorem \ref{theo1.2}.
First, we have
\begin{equation}\label{eq13.2}
\begin{split}
d_{R_{\alpha,\lambda}}(f^{\setminus i},f)+d_{R_{\alpha,\lambda}^{\setminus i}}(f,f^{\setminus i}) &= R_{\alpha,\lambda}(f^{\setminus i}) - R_{\alpha,\lambda}(f) +R_{\alpha,\lambda}^{\setminus i}(f) - R_{\alpha,\lambda}^{\setminus i}(f^{\setminus i})\\
&= \big( R_{\alpha,\lambda}(f^{\setminus i}) - R_{\alpha,\lambda}^{\setminus i}(f^{\setminus i}) \big) - \big( R_{\alpha,\lambda}(f) -R_{\alpha,\lambda}^{\setminus i}(f) \big)  \\
&= \frac 1 {mn} \sum_{j=1}^{\alpha} \mathcal L_\tau(f^{\setminus i},\pmb z_{(i,j)}) - \frac 1 {mn} \sum_{j=1}^{\alpha} \mathcal L_\tau(f,\pmb z_{(i,j)})
\end{split}
\end{equation}

Then, we have
\begin{equation}\label{eq15.2}
\begin{split}
d_{N_\lambda}(f,f^{\setminus i}) &+ d_{N_\lambda}(f^{\setminus i},f)
= d_{(R_{\alpha,\lambda}^{\setminus i} - R_\alpha^{\setminus i})}(f,f^{\setminus i}) + d_{(R_{\alpha,\lambda}-R_\alpha)}(f^{\setminus i},f)\\
&=d_{R_{\alpha,\lambda}}(f^{\setminus i},f) + d_{R_{\alpha,\lambda}^{\setminus i}}(f,f^{\setminus i}) - d_{R_\alpha}(f^{\setminus i},f) - d_{R_\alpha^{\setminus i}}(f,f^{\setminus i})\\
&\text{(based on non-negativity of Bregman divergence)}\\
&\leq d_{R_{\alpha,\lambda}}(f^{\setminus i},f) + d_{R_{\alpha,\lambda}^{\setminus i}}(f,f^{\setminus i})\\
&\text{(using (\ref{eq13.2}))}\\
&= \frac 1 {mn} \sum_{j=1}^{\alpha} \mathcal L_\tau(f^{\setminus i},\pmb z_{(i,j)}) - \frac 1 {mn} \sum_{j=1}^{\alpha} \mathcal L_\tau(f,\pmb z_{(i,j)})\\
&= \frac 1 {mn} \sum_{j=1}^{\alpha} \Bigg( \sum_{k=1}^{n/\alpha} \ell_\tau(f^{\setminus i},\pmb z_{(i,j)}, k) - \sum_{k=1}^{n/\alpha} \ell_\tau(f,\pmb z_{(i,j)}, k) \Bigg)\\
&\leq \frac 1 {mn} \sum_{j=1}^{\alpha} \sum_{k=1}^{n/\alpha} \Big| \ell_\tau(f^{\setminus i},\pmb z_{(i,j)}, k) - \ell_\tau(f,\pmb z_{(i,j)}, k) \Big|\\
&\leq \frac \tau {mn} \sum_{j=1}^{\alpha} \sum_{k=1}^{n/\alpha} \Big| f^{\setminus i}(\pmb x_{(i,j)}, k) - f(\pmb x_{(i,j)}, k) \Big|\\
&\text{(using (\ref{eq14}), and define $||\pmb x_{(i,max)}||_2 =\max_{\forall j} ||\pmb x_{(i,j)}||)_2$)}\\
&\leq \frac {\rho \tau } m   ||f-f^{\setminus i}||_2 \cdot ||\pmb x_{(i,max)}||_2\\
\end{split}
\end{equation}

This gives
\begin{equation}\label{eq17.2}
\lambda ||f-f^{\setminus i}||_2^2 \leq \frac {\rho \tau } m   ||f-f^{\setminus i}||_2 \cdot ||\pmb x_{(i,max)}||_2
\end{equation}
and thus
\begin{equation}\label{eq18.2}
||f-f^{\setminus i}||_2 \leq \frac {\rho \tau } {m\lambda} ||\pmb x_{(i,max)}||_2
\end{equation}

Then, we derive the bound of function stability $\Delta(f)$ based on sample $\pmb z$ with size $n$, and based on $\setminus i$ rather than $\setminus {i'}$. We have  $\forall \pmb z=(\pmb x, \pmb y), \forall k,$
 \begin{equation}\label{eq19.2}
\begin{split}
|f(\pmb x, k) - f^{\setminus i}(\pmb x, k)| &\leq  \rho||f- f^{\setminus i}||_2\cdot ||\pmb x||_2\\
&\text{(using (\ref{eq18.2}))}\\
&\leq \frac {\tau \rho^2 } {m\lambda}   ||\pmb x_{(i,max)}||_2 \cdot ||\pmb x||_2\\
&\leq \frac {\tau \rho^2 } {m\lambda} \cdot  \frac {nv\sqrt d} \alpha \cdot nv\sqrt d\\
&= \frac {d \tau \rho^2 v^2 n^2} {m\lambda\alpha}\\
&\text{(using (\ref{eq11}))}\\
&= \alpha \sup(\Delta) \\
\end{split}
\end{equation}
\endproof

\textbf{Proof of Theorem \ref{theo2}}

Let $f^{\setminus i}$ be defined like before. Similar to the definition of $f^{\setminus i}$ based on \emph{removing} a sample from $S$, we define $f^{i}$ based on \emph{replacing} a sample from $S$.
Let $R(f)^{\setminus i}$ denote $[R(f)]^{\setminus i}=R^{\setminus i}(f^{\setminus i})$.

First, we derive a bound for $| R(f) - R^{\setminus i}(f)|$:
\begin{equation}\label{eq1}
\begin{split}
| R(f) - R(f)^{\setminus i}|
&= \frac 1 n |\mathbb E_{\pmb z} \mathcal L_\tau(f,\pmb z) - \mathbb E_{\pmb z} \mathcal L_\tau(f^{\setminus i},\pmb z) |\\
&= \frac 1 n | \mathbb E_{\pmb z} \sum_{k=1}^n \ell_\tau (f, \pmb z, k)  -  \mathbb E_{\pmb z}\sum_{k=1}^n \ell_\tau (f^{\setminus i}, \pmb z, k) |\\
&\leq  \frac 1 n \mathbb E_{\pmb z} |  \sum_{k=1}^n \ell_\tau (f, \pmb z, k)  -  \sum_{k=1}^n \ell_\tau (f^{\setminus i}, \pmb z, k) |\\
&\leq  \frac 1 n \mathbb E_{\pmb z}  \sum_{k=1}^n |  \ell_\tau (f, \pmb z, k)  -  \ell_\tau (f^{\setminus i}, \pmb z, k) |\\
&\text{(based on Lemma \ref{lemma2} and the definition of $\bar \Delta$)}\\
&\leq  {\tau \bar\Delta} \\
\end{split}
\end{equation}

Then, we derive a bound for $| R(f) - R(f)^{i}|$:
\begin{equation*}
\begin{split}
| R(f) - R(f)^{i}| &= |R(f) - R(f)^{\setminus i}+ R(f)^{\setminus i} - R(f)^{i}| \\
&\leq |R(f) - R(f)^{\setminus i}| + |R(f)^{\setminus i} - R(f)^{i}|\\
&\text{(based on (\ref{eq1}))}\\
&\leq {\tau \bar\Delta}  + {\tau \bar\Delta} \\
&= {2\tau \bar\Delta}
\end{split}
\end{equation*}

Moreover, we derive a bound for $|R_e(f) - R_e(f)^{i}|$. Let $\pmb {\hat z}_i$ denote the full-size sample (with size $n$ and indexed by $i$) which replaces the sample $\pmb z_i$, it goes to:
\begin{equation}\label{eq2}
\begin{split}
&|R_e(f) - R_e(f)^{i}| = \Big| \frac 1 {mn} \sum_{j=1}^m \mathcal L_\tau(f,\pmb z_j) - \frac 1 {mn} \sum_{j\neq i} \mathcal L_\tau(f^{i},\pmb z_j) - \frac 1 {mn} \mathcal L_\tau(f^{i},\pmb {\hat z}_i)\Big|\\
&\leq \frac 1 {mn} \sum_{j \neq i}|\mathcal L_\tau(f, \pmb z_j) - \mathcal L_\tau(f^{i}, \pmb z_j)| + \frac 1 {mn} |\mathcal L_\tau(f, \pmb z_i) - \mathcal L_\tau(f^{i}, \pmb {\hat z}_i)|\\
&\leq \frac 1 {mn} \sum_{j \neq i}|\mathcal L_\tau(f, \pmb z_j) - \mathcal L_\tau(f^{i}, \pmb z_j)| + \frac 1 {mn} \sum_{k=1}^n  |\ell_\tau(f, \pmb z_i, k) - \ell_\tau(f^{i}, \pmb {\hat z}_i, k)|\\
&\text{(based on $0 \leq \ell_\tau (G_S, \pmb z, k) \leq \gamma$)}\\
&\leq \frac 1 {mn} \sum_{j \neq i}|\mathcal L_\tau(f, \pmb z_j) - \mathcal L_\tau(f^{i}, \pmb z_j)| + \frac {\gamma} m \\
&\leq \frac 1 {mn} \sum_{j \neq i} \Big( |\mathcal L_\tau(f, \pmb z_j) - \mathcal L_\tau(f^{\setminus i}, \pmb z_j)| + |\mathcal L_\tau(f^{\setminus i}, \pmb z_j) - \mathcal L_\tau(f^{i}, \pmb z_j)| \Big) + \frac {\gamma} m\\
&\text{(based on Lemma \ref{lemma2}, and $\Delta(f^{i},f^{\setminus i}) = \Delta(f,f^{\setminus i})$ from the definition of stability)}\\
&\leq \frac 1 {mn} \sum_{j \neq i} \Big( {n\tau \bar\Delta}  + {n\tau \bar\Delta}  \Big) + \frac {\gamma} m\\
&= \frac {2(m-1)\tau \bar\Delta+\gamma} {m} \\
\end{split}
\end{equation}

Based on the bounds of $| R(f) - R(f)^{i}|$ and $|R_e(f) - R_e(f)^{i}|$, we show that $R(f)- R_e(f)$ satisfies the conditions of \emph{McDiarmid Inequality} (Theorem \ref{theo1}) with $c_{i}=\frac {(4m-2)\tau \bar\Delta+\gamma} {m}$:
\begin{equation}\label{eq3}
\begin{split}
|[R(f)-R_e(f)] - [R(f)- R_e(f)]^{i}| &= |[R(f)-R(f)^{i}] - [R_e(f)- R_e(f)^{i}]|\\
&\leq |R(f)- R(f)^{i}| + |R_e(f) -R_e(f)^{i}|\\
&\leq {2\tau \bar\Delta}  + \frac {2(m-1)\tau \bar\Delta+\gamma} {m}\\
&=\frac {(4m-2)\tau \bar\Delta+\gamma} {m}
\end{split}
\end{equation}

Also, following the proof of Lemma \ref{lemma1}, we can get a bound for $\mathbb E_S[R(f)- R_e(f)]$:
\begin{equation}\label{eq4}
\begin{split}
\mathbb E_S[R(f) -R_e(f)]
&= \frac 1 n \mathbb E_S \Big (\mathbb E_{\pmb z}(\mathcal L (f, \pmb z))- \frac 1 m \sum_{j=1}^m \mathcal L(f, \pmb z_j) \Big ) \\
&= \frac 1 n \Big( \mathbb E_{S, \pmb {\hat z}_i} \big( \mathcal L(f, \pmb {\hat z}_i) \big) - \frac 1 m \sum_{j=1}^m \mathbb E_S \big( \mathcal L(f, \pmb z_j) \big) \Big)\\
&= \frac 1 n \Big( \mathbb E_{S, \pmb {\hat z}_i} \big( \mathcal L(f, \pmb {\hat z}_i) \big) - \mathbb E_S \big( \mathcal L(f, \pmb z_i) \big) \Big)\\
&= \frac 1 n \Big( \mathbb E_{S, \pmb {\hat z}_i} \big( \mathcal L(f, \pmb {\hat z}_i) \big) - \mathbb E_{S^i} \big( \mathcal L(f^i, \pmb {\hat z}_i) \big) \Big)\\
&= \frac 1 n \mathbb{E}_{S, \pmb {\hat z}_i} \big( \mathcal L (f, \pmb {\hat z}_i) -  \mathcal L (f^i, \pmb {\hat z}_i) \big)\\
&\leq \frac 1 n \mathbb E_{S, \pmb {\hat z}_i}|\mathcal L(f,\pmb {\hat z}_i) -\mathcal L(f^i, \pmb {\hat z}_i) |\\
&\leq \frac 1 n \mathbb E_{S, \pmb {\hat z}_i}|\mathcal L(f,\pmb {\hat z}_i) -\mathcal L(f^{\setminus i}, \pmb {\hat z}_i) | + \frac 1 n \mathbb E_{S, \pmb {\hat z}_i}|\mathcal L(f^{\setminus i},\pmb {\hat z}_i) -\mathcal L(f^i, \pmb {\hat z}_i) |\\
& \text{(based on Lemma \ref{lemma2} and the $\bar \Delta$ defined in (\ref{eq11.2}))}\\
&\leq {\tau \bar \Delta}  + {\tau \bar \Delta} \\
&= {2\tau \bar \Delta}
\end{split}
\end{equation}

Now, we can apply \emph{McDiarmid Inequality} (Theorem \ref{theo1}):
\begin{equation}
\mathbb P_S \Big( [R(f) -R_e(f)] -\mathbb E_S[R(f) -R_e(f)] \geq \epsilon \Big) \leq \exp{\Big( \frac {-2\epsilon^2} {\sum_{i=1}^{m} c_{i}^2} \Big)}
\end{equation}
Based on (\ref{eq3}) and (\ref{eq4}), it goes to
\begin{equation}\label{eq5}
\mathbb P_S \Big( R(f) -R_e(f) \geq {2\tau \bar\Delta}  + \epsilon \Big) \leq \exp{\Bigg( \frac {-2m \epsilon^2} {\big((4m-2)\tau \bar\Delta+\gamma \big)^2} \Bigg)}
\end{equation}
Let $\delta=  \exp{\Big( \frac {-2m \epsilon^2} {\big((4m-2)\tau \bar\Delta +\gamma \big)^2} \Big)}$, we have
\begin{equation}\label{eq6}
\epsilon = \Big( (4m-2)\tau \bar\Delta  + \gamma \Big) \sqrt{\frac {\ln {\delta^{-1}}} {2m}}
\end{equation}

Based on (\ref{eq5}) and (\ref{eq6}), there is a probability no more than $\delta$ such that
\begin{equation}
\begin{split}
R(f) -R_e(f) &\geq {2\tau \bar\Delta}  + \epsilon \\
&= {2\tau \bar\Delta}  + \Big({(4m-2)\tau \bar\Delta}  + \gamma \Big) \sqrt{\frac {\ln {\delta^{-1}}} {2m}}\\
\end{split}
\end{equation}

Then, there is a probability at least $1-\delta$ such that
$$
R(f) \leq R_e(f) + {2\tau \bar\Delta}  + \Big({(4m-2)\tau \bar\Delta}  + \gamma \Big) \sqrt{\frac {\ln {\delta^{-1}}} {2m}}
$$
which gives (\ref{eq10}).
\endproof

\textbf{Proof of Theorem \ref{theo3}}

According to (\ref{eq11.2}), we have $\bar\Delta \leq \frac {d \tau \rho^2 v^2 n^2} {m\lambda\alpha}$. Inserting this bound into (\ref{eq10}) gives (\ref{eq23}).
\endproof

\textbf{Proof of Proposition \ref{theo4}}

By subtracting $\pmb w^*$ from both sides and taking norms for (\ref{eq29}), we have
\begin{equation}
\begin{split}
||\pmb w_{t+1}-\pmb w^*||^2 &= ||\pmb w_t - \eta \nabla g_{\pmb z_t}(\pmb w_t) - \pmb w^*||^2\\
&= ||\pmb w_t - \pmb w^*||^2 -2\eta(\pmb w_t -\pmb w^*)^T \nabla g_{\pmb z_t}(\pmb w_t) + \eta^2||\nabla g_{\pmb z_t}(\pmb w_t)||^2
\end{split}
\end{equation}
Taking expectations and let $a_t=||\pmb w_t - \pmb w^*||^2$, we have
\begin{equation}\label{eq34}
\begin{split}
a_{t+1}
&= a_t -2\eta \mathbb E [(\pmb w_t -\pmb w^*)^T \nabla g_{\pmb z_t}(\pmb w_t)] + \eta^2 \mathbb E [||\nabla g_{\pmb z_t}(\pmb w_t)||^2]\\
&\text{(based on (\ref{eq32}) )}\\
&\leq a_t -2\eta \mathbb E [(\pmb w_t -\pmb w^*)^T \nabla g_{\pmb z_t}(\pmb w_t)] + {\eta^2 \kappa^2 |\pmb z_t|^2} \\
&\text{(Recall $\pmb z_t$ is of the size $n/\alpha$ based on the definition of structure regularization )}\\
&= a_t -2\eta \mathbb E [(\pmb w_t -\pmb w^*)^T \nabla g_{\pmb z_t}(\pmb w_t)] + \frac {\eta^2 \kappa^2 n^2} {\alpha^2} \\
&\text{(since the random drawing of $\pmb z_t$ is independent of $\pmb w_t$)}\\
&= a_t -2\eta \mathbb E [(\pmb w_t -\pmb w^*)^T \mathbb E_{\pmb z_t}(\nabla g_{\pmb z_t}(\pmb w_t))] + \frac {\eta^2 \kappa^2 n^2} {\alpha^2} \\
&= a_t -2\eta \mathbb E [(\pmb w_t -\pmb w^*)^T \nabla g(\pmb w_t)] + \frac {\eta^2 \kappa^2 n^2} {\alpha^2} \\
\end{split}
\end{equation}
By setting $\pmb w' = \pmb w^*$ in (\ref{eq30}), we have
\begin{equation}\label{eq35}
\begin{split}
(\pmb w - \pmb w^*)^T \nabla g(\pmb w)
&\geq g(\pmb w) - g(\pmb w^*) + \frac c 2 ||\pmb w - \pmb w^*||^2\\
&\geq \frac c 2 ||\pmb w - \pmb w^*||^2
\end{split}
\end{equation}
Combining (\ref{eq34}) and (\ref{eq35}), we have
\begin{equation}\label{eq36}
\begin{split}
a_{t+1} &\leq a_t -\eta c ||\pmb w_t - \pmb w^*||^2  + \frac {\eta^2 \kappa^2 n^2} {\alpha^2}\\
&= (1-c\eta)a_t + \frac {\eta^2 \kappa^2 n^2} {\alpha^2}
\end{split}
\end{equation}
We can find the steady state $a_\infty$ as follows
\begin{equation}
a_\infty = (1-c\eta)a_\infty + \frac {\eta^2 \kappa^2 n^2} {\alpha^2}
\end{equation}
which gives
\begin{equation}\label{eq40}
a_\infty = \frac {\eta \kappa^2 n^2} {c \alpha^2}
\end{equation}

Defining the function $A(x)=(1-c\eta)x + \frac {\eta^2 \kappa^2 n^2} {\alpha^2}$, based on (\ref{eq36}) we have
\begin{equation}\label{eq37}
\begin{split}
a_{t+1} &\leq A(a_t) \\
&\text{(Taylor expansion of $A(\cdot)$ based on $a_\infty$, with $\nabla^2 A(\cdot)$ being 0)}\\
&= A(a_\infty)+ \nabla A(a_\infty)(a_t-a_\infty)\\
&= A(a_\infty) + (1-c\eta)(a_t-a_\infty)\\
&= a_\infty + (1-c\eta)(a_t-a_\infty)
\end{split}
\end{equation}
Unwrapping (\ref{eq37}) goes to
\begin{equation}\label{eq38}
a_t <= (1-c\eta)^t (a_0 - a_\infty) + a_\infty
\end{equation}

Since $\nabla g(\pmb w)$ is Lipschitz according to (\ref{eq31}), we have
$$
g(\pmb w) \leq g(\pmb w') + \nabla g(\pmb w')^T (\pmb w - \pmb w') + \frac q 2 ||\pmb w- \pmb w'||^2
$$
Setting $\pmb w' = \pmb w^*$, it goes to $g(\pmb w) - g(\pmb w^*) \leq \frac q 2 ||\pmb w- \pmb w^*||^2$, such that
$$
\mathbb E [g(\pmb w_t)- g(\pmb w^*)] \leq \frac q 2 ||\pmb w_t- \pmb w^*||^2 = \frac q 2 a_t
$$
In order to have $E [g(\pmb w_t)- g(\pmb w^*)] \leq \epsilon$, it is required that $\frac q 2 a_t \leq \epsilon$, that is
\begin{equation}\label{eq39}
a_t \leq \frac {2\epsilon} q
\end{equation}

Combining (\ref{eq38}) and (\ref{eq39}), it is required that
\begin{equation}
(1-c\eta)^t (a_0 - a_\infty) + a_\infty \leq \frac {2\epsilon} q
\end{equation}
To meet this requirement, it is sufficient to set the learning rate $\eta$ such that both terms on the left side are less than  $\frac \epsilon q$. For the requirement of the second term $a_\infty \leq \frac \epsilon q$, recalling (\ref{eq40}), it goes to
$$
\eta \leq \frac {c\epsilon \alpha^2} {q \kappa^2 n^2}
$$
Thus, introducing a real value $\beta \in (0,1]$, we can set $\eta$ as
\begin{equation}\label{eq41}
\eta = \frac {c\epsilon \beta \alpha^2} {q \kappa^2 n^2}
\end{equation}

On the other hand, for the requirement of the first term $(1-c\eta)^t (a_0 - a_\infty) \leq \frac \epsilon q$, it goes to
\begin{equation}\label{eq42}
\begin{split}
t &\geq \frac {\log {\frac \epsilon {q a_0}}} {\log {(1-c\eta)}}\\
&\text{(since $\log {(1-c\eta)} \leq -c\eta$ given (\ref{eq33}))} \\
&\geq \frac {\log {(q a_0 / \epsilon)}} {c\eta}
\end{split}
\end{equation}
Combining (\ref{eq41}) and (\ref{eq42}), it goes to
$$
t \geq
\frac {q \kappa^2 n^2 \log {(q a_0 / \epsilon)}} {\epsilon \beta c^2 \alpha^2}
$$
which completes the proof.
\endproof

\section{Conclusions}
We proposed a structure regularization framework, which decomposes training samples into mini-samples with simpler structures, deriving a trained model with regularized structural complexity. Our theoretical analysis showed that this method can effectively reduce the generalization risk, and can also accelerate the convergence speed in training.
The proposed method does not change the convexity of the objective function, and can be used together with any existing weight regularization methods.
Note that, the proposed method and the theoretical results can fit general structures including linear chains, trees, and graphs.
Experimental results demonstrated that our method achieved better results than state-of-the-art systems on several highly-competitive tasks, and at the same time with substantially faster training speed.

The structure decomposition of structure regularization can naturally used for parallel training, achieving parallel training over each single samples. As future work, we will combine structure regularization with parallel training.

See \cite{Sun_NIPS2014} for the conference version of this work.

\subsubsection*{Acknowledgments}
This work was supported in part by National
Natural Science Foundation of China
(No. 61300063), and Doctoral
Fund of Ministry of Education of China
(No. 20130001120004).

\small
\bibliographystyle{abbrv}
\bibliography{bib}

\end{document}